\documentclass[twocolumn]{article}
\makeatletter

\usepackage{tabularx}
\usepackage{ragged2e}
\usepackage{bm}
\usepackage{boldline}

\newcolumntype{C}{>{\Centering\arraybackslash}X}

\newcolumntype{u}{>{\raggedright\hsize=.7\hsize}X}
\newcolumntype{t}{>{\Centering\hsize=.6\hsize}X}
\newcolumntype{s}{>{\Centering\hsize=.5\hsize}X}
\newcolumntype{o}{>{\Centering\hsize=.4\hsize}X}
\newcolumntype{k}{>{\Centering\hsize=.3\hsize}X}
\newcolumntype{y}{>{\Centering\hsize=.2\hsize}X}
\newcolumntype{z}{>{\Centering\hsize=.1\hsize}X}
\newcolumntype{v}{>{\raggedright\hsize=.6\hsize}X}
\newcolumntype{e}{>{\raggedright\hsize=.5\hsize}X}
\newcolumntype{j}{>{\raggedright\hsize=.35\hsize}X}
\newcolumntype{f}{>{\raggedright\hsize=.3\hsize}X}
\newcolumntype{h}{>{\raggedright\hsize=.2\hsize}X}
\newcolumntype{q}{>{\raggedright\hsize=.8\hsize}X}

\usepackage{tabulary,graphicx,amsmath,amsfonts,amssymb,setspace,caption}
\usepackage{booktabs}
\usepackage{multirow}
\usepackage{float}
\usepackage{subfig}
\usepackage{graphicx}
\usepackage[margin=0.7in]{geometry}

\usepackage{subfig}
\usepackage[switch]{lineno}
\usepackage{color}

\usepackage{zref-xr,zref-user} 

\usepackage{url}

\def\x{{\mathbf x}}

\def\y{{\mathbf y}}

\usepackage{pifont}
\newcommand{\cmark}{\ding{51}}%
\newcommand{\xmark}{\ding{55}}%

\newenvironment{affiliations}{%
    \setcounter{enumi}{1}%
    \setlength{\parindent}{0in}%
    \slshape\sloppy
    \begin{list}{\upshape$^{\arabic{enumi}}$}{%
        \usecounter{enumi}%
        \setlength{\leftmargin}{0in}%
        \setlength{\topsep}{0in}%
        \setlength{\labelsep}{0in}%
        \setlength{\labelwidth}{0in}%
        \setlength{\listparindent}{0in}%
        \setlength{\itemsep}{0ex}%
        \setlength{\parsep}{0in}%
        }
    }{\end{list}\par\vspace{12pt}}

\renewenvironment{abstract}{%
    \textbf{Abstract}---\setlength{\parindent}{0in}%
    \setlength{\parskip}{0in}%
    }{\par\vspace{25pt}}
\makeatother

\makeatother
\usepackage[normalem]{ulem}
\usepackage{xcolor}
\definecolor{nmiblue}{HTML}{0063EA}

\usepackage{hyperref}

\begin{document}

\title{Visual Speech Recognition for Multiple Languages in the Wild}
\author{Pingchuan Ma\(^{1}\)\hspace{.2pc},
        Stavros Petridis\(^{1,2}\)\hspace{.2pc},
        Maja Pantic\(^{1,2}\)
    }
\date{}

\twocolumn[
  \begin{@twocolumnfalse}
    \maketitle
    \begin{affiliations}
      \item
        Imperial College London, London, UK
      \item Meta AI, London, UK
    \end{affiliations}
    \begin{abstract}
Visual speech recognition (VSR) aims to recognize the content of speech based on lip movements, without relying on the audio stream. Advances in deep learning and the availability of large audio-visual datasets have led to the development of much more accurate and robust VSR models than ever before. However, these advances are usually due to the larger training sets rather than the model design. Here we demonstrate that designing better models is equally as important as using larger training sets. We propose the addition of prediction-based auxiliary tasks to a VSR model, and highlight the importance of hyperparameter optimization and appropriate data augmentations. We show that such a model works for different languages and outperforms all previous methods trained on publicly available datasets by a large margin. It even outperforms models that were trained on non-publicly available datasets containing up to to 21 times more data. We show, furthermore, that using additional training data, even in other languages or with automatically generated transcriptions, results in further improvement.    \end{abstract}
  \end{@twocolumnfalse}
]

\section{Introduction}
Visual speech recognition (VSR), also known as lipreading, is the task of automatically recognizing speech from video based only on lip movements. In the past, this field has attracted a lot of research attention within the speech recognition community~\cite{DBLP:journals/pieee/PotamianosNGGS03,DBLP:journals/tmm/DupontL00} but it has failed to meet the initial high expectations. There are two main reasons why the first generation of VSR models fell short: (1) the lack of large transcribed audio-visual datasets resulted in models that could only recognize a limited vocabulary and work only in a laboratory environment and (2) the use of handcrafted visual features, which might not have been optimal for VSR applications, prevented the development of high-accuracy models. Recently, large audio-visual transcribed datasets, like LRS2~\cite{chung2017lip} and LRS3~\cite{afouras2018deep}, have become available, and these have allowed the development of a large vocabulary and robust models. In addition, advances in deep learning have made possible the use of end-to-end models, which learn to extract VSR-related features directly from raw images. These developments have led to a new generation of deep-learning-based VSR models that achieve much higher accuracy than older models and also work in unseen real-life situations.

The recent advances in VSR models are mainly fuelled by using increasingly larger transcribed datasets and the development of models that work well when trained with huge amounts of data. Some recent works~\cite{shillingford2019large, serdyuk2021audiovisual} use tens of thousands of hours of non-publicly available training data to achieve state-of-the-art performance on standard benchmarks. In contrast to this recent trend, we demonstrate that carefully designing a model is equally as important as using larger training sets. Our approach revolves around (1) addition of prediction-based auxiliary tasks to a VSR model, (2) appropriate data augmentations and (3) hyperparameter optimization of an existing architecture. This leads to a great reduction in word error rate (WER) and results in state-of-the-art performance on almost all benchmarks. This is achieved by using only publicly available datasets, which are two orders of magnitude smaller than those used in previous works. We also show that combining multiple datasets further improves the performance (which is in line with the results reported in the literature). Hence, we argue that further progress in the field can be achieved not only by increasing the size of the training data but also by careful model design and optimization.       

The vast majority of existing works focus on improving the performance of English-only VSR models. There are also a few works that design models tailored to a specific language, like Mandarin~\cite{DBLP:conf/aaai/ZhangGDYL019, zhao2019cascade,ma2020transformer}. In contrast to previous works, our approach is evaluated not only on English but also on Mandarin and Spanish (the two other widely spoken languages), Italian, French and Portuguese. State-of-the-art performance is achieved in all languages.

Specifically, in this Article, we make the following contributions:
\begin{itemize}

    \item We propose a novel method for VSR that outperforms state-of-the-art methods trained on publicly available data by a large margin.

    \item We do so with a VSR model with auxiliary tasks that jointly performs VSR and prediction of audio and visual representations.

    \item We demonstrate that the proposed VSR model performs well, not only in English, but also in other languages, such as Spanish, Mandarin, Italian, French and Portuguese.
    
    \item We show that enlarging the training sets, even by including unlabelled data with automatically generated transcriptions or videos in other languages, results in improved performance. This provides further evidence for the hypothesis that the recent improvements presented in the literature are probably the result of larger training sets and not necessarily of better models.

    \item We discuss challenges for VSR systems that need to be solved and ethical considerations that must be taken into account before this technology can be widely applied.

\end{itemize}

\subsection{Baseline VSR Model}
\label{ssec:VSR_model}
The baseline VSR model that we extend in this work is based on~\cite{DBLP:journals/corr/abs-2102-06657}. The model consists of a three-dimensional~(3D) convolutional layer with a receptive field of five frames, followed by a 2D ResNet-18 (Fig.~\ref{fig:av_architecture}e)
, a 12-layer Conformer model~\cite{gulati2020conformer} and a transformer decoder as shown in Fig.~\ref{fig:av_architecture}b. The model is trained end to end using a combination of the connectionist temporal classification~(CTC) loss with an attention mechanism. Data augmentation is also used during training in the form of random cropping and image flipping (applied to all frames in the same sequence). This model achieves state-of-the-art VSR performance on the LRS2 and LRS3 datasets, when only publicly available data are used for training.

\subsection{Baseline ASR Model}
\label{ssec:ASR_model}
\begin{figure*}[tb]
\includegraphics{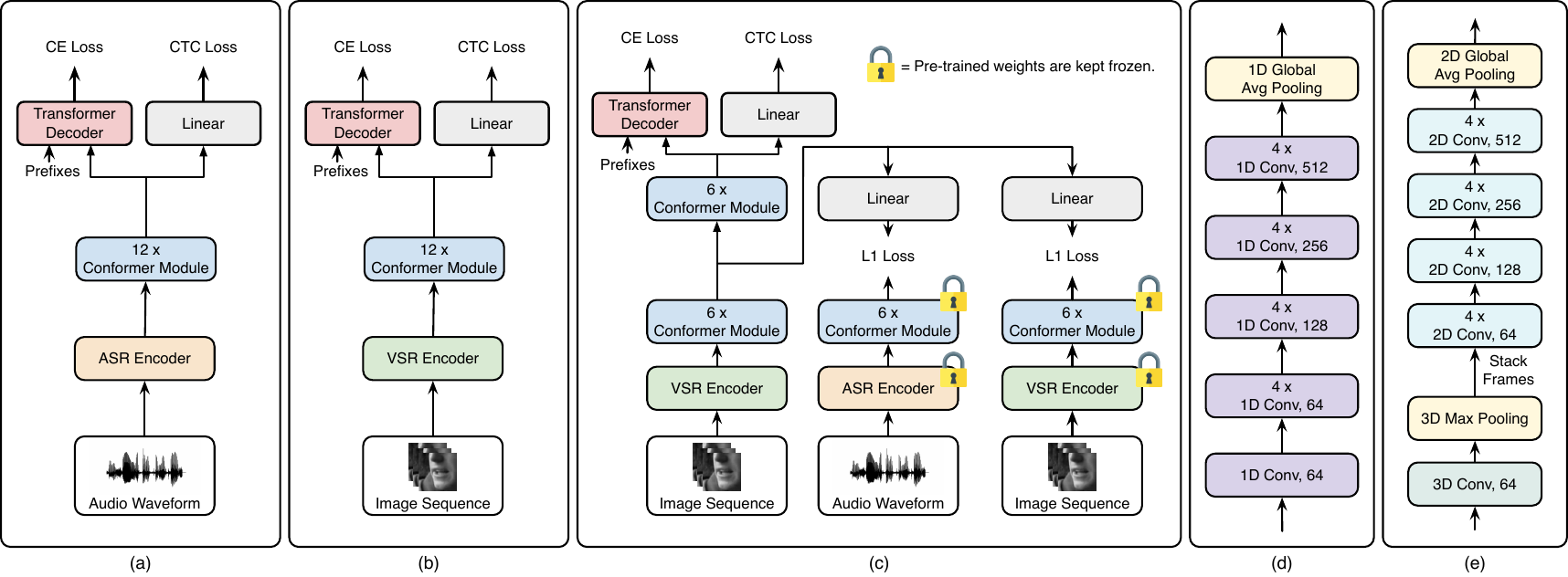}
\caption{Model architecture overview. a-c, Baseline ASR model (a), baseline VSR model (b) and proposed model (c) with prediction-based auxiliary tasks. The frame rate of extracted visual features and audio features is 25. (d), The architecture of the ASR encoder from a. e, The architecture of the VSR encoder from b.}
\label{fig:av_architecture}
\end{figure*}

The baseline Automatic Speech Recognition (ASR) model that we use is based on~\cite{DBLP:journals/corr/abs-2102-06657}. The model consists of an 1D ResNet-18 (Fig.~\ref{fig:av_architecture}d), a 12-layer Conformer model and a transformer decoder as shown in Fig.~\ref{fig:av_architecture}a. This model also follows the hybrid CTC/attention architecture and is trained end to end. Time-masking is also used as data augmentation during training. At the moment, this is the state-of-the-art ASR model on the LRS2 and LRS3 datasets.

\subsection{Our Approach}
In contrast to previous works, which improve the VSR performance by using increasingly larger training sets, we focus on improving the performance by carefully designing a model without relying on additional data. This is achieved by revising the training strategy and architecture of the state-of-the-art model proposed in~\cite{DBLP:journals/corr/abs-2102-06657}.
First, we optimize hyperparameters and improve the language model (LM) with the aim of squeezing extra performance out of the model. Second, we introduce time-masking, which is a temporal augmentation method that is commonly used in ASR models. It substantially improves the VSR performance by forcing the model to rely more on contextual information and, as a consequence, it can better disambiguate similar lip movements that correspond to different phonemes. Finally, we use a VSR model with auxiliary tasks where the model jointly performs VSR and prediction of audio and visual representations extracted from pre-trained VSR and ASR models. This prediction task provides an additional supervisory signal and forces the model to learn better visual representations. A diagram of the architecture of our model is shown in Fig.~\ref{fig:av_architecture}c.

The performance of our model is presented in Tables~\ref{table: results_on_LRS2} to~\ref{tab: cmumoseas_results}. Owing to the random nature of training, we train ten models for each experiment and report the mean and standard deviation of the WER over the ten runs. This is in contrast to previous works, which report just a single value (most probably the best WER) and no standard deviation, and it provides a more robust estimate of the actual performance. However, to facilitate a fair comparison with other works, we also report the best WER of the ten runs.

\subsection{Results on LRS2}\label{ssec:results_on_lrs2}
\begin{table*}[!t]
\caption{Results on the LRS2 dataset. `Mean$\pm$Std.' refers to the mean word error rate over ten runs and the corresponding standard deviation, while ``Best'' denotes the best (lowest) WER.}
\renewcommand\arraystretch{1.1}
\begin{tabularx}{2\columnwidth}{j v k o y y}
\toprule
Method &Pre-training Set &Training Set &Training Sets Total Size (hours) &Mean$\pm$Std. &Best \\
\midrule\midrule
\multicolumn{6}{c}{\textit{Using Publicly Available Datasets}} \\ 
\midrule
MV-WAS~\cite{chung2017lip} &-  &LRS2 &223 &- &70.4 \\
\midrule
CTC/Att.~\cite{petridis2018audio} &LRW &LRS2 &380  &- &63.5 \\
\midrule
KD\,+\,CTC~\cite{afouras2020asr} &VoxCeleb2$^{\text{clean}}$+LRS3 &LRS2 &995 &- &51.3 \\
\midrule
KD-seq2seq~\cite{Ren_2021_CVPR} &LRW+LRS3 &LRS2 &818 &- &49.2 \\
\midrule
TDNN~\cite{yu2020audio} &- &LRS2 &223 &- &48.9 \\
\midrule
CM-seq2seq~\cite{DBLP:journals/corr/abs-2102-06657} &LRW &LRS2 &380 &- &37.9 \\
\midrule
Ours &- &LRS2 &223 &\bf 33.6$\pm$0.5 &\bf 32.9 \\
\midrule
Ours &LRW &LRS2 &380 &\bf 29.5$\pm$0.4 &\bf 28.7  \\
\midrule
Ours &LRW+LRS3 &LRS2 &818 &\bf 27.6$\pm$0.2 &\bf 27.3  \\
\midrule
Ours &LRW+LRS3+AVSpeech &LRS2 &1\,459 &\bf 25.8$\pm$0.4 &\bf 25.5   \\
\midrule
\multicolumn{6}{c}{\textit{Using Non-Publicly Available Datasets}} \\
\midrule
TM-seq2seq~\cite{afouras2018deep} &MVLRS+LRS3 &LRS2 &1\,391  &- &48.3  \\
\bottomrule
\end{tabularx}
\label{table: results_on_LRS2}
\end{table*}

The results on LRS2—an English audio-visual dataset—are reported in Table~\ref{table: results_on_LRS2}. Our model outperforms all existing works by a large margin, even when it is trained on smaller amounts of training data. In particular, it outperforms the previous state of the art~\cite{DBLP:journals/corr/abs-2102-06657}, in terms of the best WER achieved, by 5\,\%. This is despite the fact that~\cite{DBLP:journals/corr/abs-2102-06657} is trained on a larger training set. When we use the same training set size as in~\cite{DBLP:journals/corr/abs-2102-06657} our model results in a 9.2\,\% improvement. When we use additional training data, an even larger improvement of 12.4\,\% is observed. Similarly, our approach results in a 22.8\,\% absolute improvement in the best WER over~\cite{afouras2018deep} which uses a training set with similar size to ours and also includes non-publicly available data.

\subsection{Results on LRS3}\label{ssec:results_on_lrs3}
\begin{table*}[!t]
\renewcommand\arraystretch{1.1}
\caption{Results on the LRS3 dataset. `Mean$\pm$Std.' refers to the mean word error rate over ten runs and the corresponding standard deviation, while ``Best'' denotes the best (lowest) WER.}
\begin{tabularx}{2\columnwidth}{j v k o y y}
\toprule
Method &Pre-training Set &Training Set &Training Sets Total Size (hours) &Mean$\pm$Std. &Best \\
\midrule\midrule
\multicolumn{6}{c}{\textit{Using Publicly Available Datasets}} \\ \midrule
KD+CTC~\cite{afouras2020asr} &VoxCeleb2$^{\text{clean}}$ &LRS3 &772 &- &59.8\\\midrule
KD-seq2seq~\cite{Ren_2021_CVPR} &LRW+LRS2 &LRS3 &818 &- &59.0 \\ \midrule
CM-seq2seq~\cite{DBLP:journals/corr/abs-2102-06657} &LRW &LRS3 &595 &- &43.3 \\\midrule

Ours &- &LRS3 &438 &\bf 38.6$\pm$0.4 &\bf 37.9 \\ \midrule

Ours &LRW &LRS3 &595 &\bf 35.8$\pm$0.5 &\bf 35.1 \\ \midrule

Ours &LRW+LRS2 &LRS3 &818 &\bf 34.9$\pm$0.2 &\bf 34.7 \\ \midrule

Ours &LRW+LRS2+AVSpeech &LRS3 &1\,459 &\bf 32.1$\pm$0.3 &\bf 31.5 \\ \midrule
\multicolumn{6}{c}{\textit{Using Non-Publicly Available Datasets}} \\ \midrule
TM-seq2seq~\cite{afouras2018deep} &MVLRS+LRS2 &LRS3 &1\,391 &- &58.9  \\\midrule
V2P~\cite{shillingford2019large} &-  &LSVSR &3\,886 &- &55.1 \\\midrule
RNN-T~\cite{makino2019recurrent} &- &YT-31k &31\,000 &- &33.6 \\\midrule
ViT3D-TM~\cite{serdyuk2021audiovisual} &- &YT-90k &90\,000 &- &25.9 \\ \midrule
ViT3D-CM~\cite{serdyuk2022transformer} &- &YT-90k &90\,000 &- &17.0 \\ 
\bottomrule
\end{tabularx}
\label{table: results_on_LRS3}
\end{table*}
The results on LRS3—another English audio-visual dataset—are presented in Table~\ref{table: results_on_LRS3}. In this case too, our proposed approach substantially outperforms all existing works that are trained using publicly available datasets. In particular, our method leads to an 8.2\,\% absolute improvement, in terms of the best WER, over the state of the art~\cite{DBLP:journals/corr/abs-2102-06657} when the same training data are used. As expected, a smaller absolute improvement of 5.4\,\% is reported when a smaller training set is used. In the case of additional training data being available, a larger absolute improvement of 11.8\,\% is achieved.

There are also some works that rely on very large non-publicly available datasets for training. As a consequence, it is not clear whether the reported improvement in WER is due to a better model or simply to the large amount of training data. Our approach outperforms all works that use up to 21 times more training data. More specifically, our best model, trained on 1\,453 h of video, leads to a 2.1\,\% absolute improvement over~\cite{makino2019recurrent} which uses 31\,000 hours of training data. However, it performs worse than~\cite{serdyuk2021audiovisual}, which presents a model trained on 90\,000 hours, which is 62 times more training data than the publicly available training data on which our model is trained.

\subsection{Results on CMLR}\label{ssec:results_on_cmlr}
\begin{table*}[!t]
\centering
\caption{Results on the CMLR dataset. `Mean$\pm$Std.' refers to the mean character error rate over ten runs and the corresponding standard deviation, while ``Best'' denotes the best (lowest) CER.}
\renewcommand\arraystretch{1.1}
\begin{tabularx}{2\columnwidth}{j v k o y y}
\toprule
Method &Pre-training Set &Training Set &Training Sets Total Size (hours) &Mean$\pm$Std. &Best \\
\midrule\midrule
    LipCH-Net \cite{DBLP:conf/aaai/ZhangGDYL019} & - &CMLR &61 & &34.0 \\
\midrule
    CSSMCM \cite{zhao2019cascade} & - &CMLR &61   &- &32.5 \\
\midrule
    LIBS \cite{zhao2020hearing} &- &CMLR &61  &- &31.3 \\
\midrule
    CTCH \cite{ma2020transformer}  &-  &CMLR &61  &- &22.0 \\
\midrule

    Ours &- &CMLR &61  &\bf 9.1$\pm$0.05 &\bf 9.1 \\
\midrule

    Ours &LRW+LRS2+LRS3 &CMLR &879  &\bf 8.2$\pm$0.06 &\bf 8.1 \\

\midrule

    Ours &LRW+LRS2+LRS3+AVSpeech &CMLR &1\,520 &\bf 8.1$\pm$0.05 &\bf 8.0 \\
\bottomrule
\end{tabularx}
\label{tab: cmlr_results}
\end{table*}
The results on the CMLR dataset—a Mandarin audio-visual dataset—are shown in Table~\ref{tab: cmlr_results}. We report performance in terms of character error rate (CER) instead of WER, because Chinese characters are not separated by spaces. Our approach results in a substantial reduction in the CER over all existing works. We achieve an absolute improvement of 12.9\,\% over the state of the art~\cite{ma2020transformer}. The WER can be further reduced by 1.1\,\% by first pre-training our model on English and then fine-tuning it on the CMLR training set.

\subsection{Results on CMU-MOSEAS-Spanish}\label{ssec:results_on_spanish}
\begin{table*}[!t]
\centering
\caption{Results on the CMU-MOSEAS-Spanish (CM$_{\text{es}}$) dataset. `Mean$\pm$Std.' refers to the mean word error rate over ten runs and the corresponding standard deviation, while ``Best'' denotes the best (lowest) WER.}
\renewcommand\arraystretch{1.1}
\begin{tabularx}{2\columnwidth}{j v k o y y}
\toprule
Method &Pre-training Set &Training Set &Training Sets Total Size (hours) &Mean$\pm$Std. &Best \\
\midrule

CM-seq2seq~\cite{DBLP:journals/corr/abs-2102-06657} &LRW &CM$_{\text{es}}$+MT$_{\text{es}}$ &244 & 58.9$\pm$0.8 & 58.1 \\ \midrule

Ours &LRW &CM$_{\text{es}}$+MT$_{\text{es}}$ &244 &\bf 51.5$\pm$0.8 &\bf 50.4 \\ \midrule

Ours &LRW+LRS2+LRS3 &CM$_{\text{es}}$+MT$_{\text{es}}$ &905 &\bf 47.4$\pm$0.2 &\bf 47.2 \\ \midrule

Ours &LRW+LRS2+LRS3+AVSpeech &CM$_{\text{es}}$+MT$_{\text{es}}$ &1\,546 &\bf 44.6$\pm$0.6 &\bf 43.9 \\

\bottomrule
\end{tabularx}
\label{tab: cmumoseas_results}
\end{table*}
The results on the CMU-MOSEAS-Spanish dataset—an audio-visual Spanish dataset—are shown in Table~\ref{tab: cmumoseas_results}. Given that this is a small dataset, it is not possible to train an accurate model without using additional data. For this purpose, we first pre-trained the model on English datasets and then fine-tuned it on the training sets of CMU-MOSEAS and TEDx datasets using the Spanish videos only. Because this is a new dataset and there are no results from previous works, we trained the end-to-end model presented in~\cite{DBLP:journals/corr/abs-2102-06657} to serve as the baseline. We observe that our proposed approach results in a 7.7\,\% absolute reduction in the WER. A further reduction of 6.5\,\% can be achieved by using additional training data.

\subsection{Comparison between Mean and Best WER/CER}
In all results shown in Tables~\ref{table: results_on_LRS2} to~\ref{tab: cmumoseas_results} we report both the mean and the best performance over ten runs. We observe that the mean WER, which is more representative of the actual performance, is up to 0.8\,\% worse than the best WER. The only exception is for the CMLR dataset (Table~\ref{tab: cmlr_results}), where the mean and best CER are practically the same, mainly as a result of the large size of the test set. This difference between the mean and best WER is something that should be taken into account when comparing different models, especially when the models are tested on relatively small test sets and the results are too close.

\section{Applications, Challenges and Ethical Considerations}
\subsection{Applications}
Speech is the most commonly used human communication method and consists of an audio signal and the corresponding mouth movements. Speech perception is also bimodal, as demonstrated by the McGurk effect~\cite{mcgurk1976hearing}, where the perception of a sound may change depending on the lip movements shown to the observers. In addition, it has been shown that the addition of visual speech information to a word recognition task performed by normal hearing adults is equivalent to increasing the signal-to-noise ratio (SNR) by $15$\,dB compared to audio-only recognition~\cite{sumby1954visual}. Hence, one of the main applications of VSR is to enhance the performance of ASR models in noisy environments. VSR models are not substantially affected by acoustic noise and can be integrated into an audio-visual speech recognition (AVSR) model to compensate for the performance drop of ASR models. Several AVSR architectures have been proposed~\cite{afouras2018deep, DBLP:journals/corr/abs-2102-06657, makino2019recurrent, petridis2018audio, yu2020audio, DBLP:conf/icassp/YuZK21,DBLP:journals/taslp/SterpuSH20}; these show that the improvement over ASR models is greater as the noise level increases, that is, the SNR is lower. The same VSR architectures can also be used to improve the performance of audio-based models in a variety of applications like speech enhancement~\cite{DBLP:conf/interspeech/AfourasCZ18}, speech separation~\cite{DBLP:journals/tog/EphratMLDWHFR18}, voice activity detection~\cite{DBLP:conf/icassp/YoshimuraHT020}, active speaker detection~\cite{DBLP:journals/corr/abs-2108-07640} and speaker diarization~\cite{DBLP:conf/interspeech/ChungHNAZ20}.

There are also a number of applications based exclusively on VSR. Silent speech interfaces (SSIs)~\cite{denby2010silent}, which can enable speech communication to take place when an audible speech signal is not available, can be developed with the help of VSR systems. This means that a speaker would be able to mouth words instead of vocalizing them. This technology has the potential to transform the lives of speech-impaired people. Individuals who have lost the ability to speak (aphonia) or have difficulty in speaking (dysphonia) due to tracheostomy, laryngectomy, stroke or injury might find it hard to communicate with others. The use of SSI can alleviate this by providing an alternative way of communication and at the same time reduce the stress caused by the sudden loss of their voice. The use of SSI can also be useful in cases where speaking is not allowed, for example, in a meeting, and can provide privacy in public conversations.

VSR technology also opens up opportunities to automatically transcribe video content that was recorded without audio, like silent movies, CCTV footage or video captured by older webcams, and would otherwise require substantial manual effort or might have even been impossible. It can also be used as a useful tool in face forgery detection~\cite{haliassos2021lips}. Most face-manipulation approaches add inconsistencies in mouth movements, which might not always be perceptible by humans, but they can easily be detected by properly trained VSR models. Finally, there is a new form of VSR that has become popular recently and generates audio, instead of text, directly from the input video~\cite{mira2021end,prajwal2020learning}. This is essentially a combination of a standard VSR model with a text-to-speech model, but it has two important advantages: (1) it does not require any transcribed dataset and can be trained with vast amounts of unlabelled audio-visual data, and (2) it is faster and can potentially be used in real-time applications as it removes the constraint of recognizing a complete word before generating the corresponding speech signal. This new approach is especially useful for audio inpainting applications, because it can automatically fill in audio gaps from video.

\subsection{Challenges}
Despite the great advances in VSR, there are still numerous challenges that need to be solved before the full potential of this technology can be achieved. First, visual ambiguities that arise from the fact that different phonemes correspond to similar lip movements is one of the most important reasons for the substantial performance gap between ASR and VSR models. Designing VSR systems that can resolve some of these ambiguities by relying more on the context, like the time-masking augmentation proposed in this work, might close this gap. In addition, VSR systems are sensitive to visual noise like lighting changes, occlusions, motion blur and compression. Reduced and/or mismatched resolution and frame rate between training and test conditions can also affect performance. There is some evidence that VSR systems are robust to small or moderate amounts of noise and less robust to reduced resolution~\cite{Dungan2018,bear2014resolution}, but further studies are needed to establish the impact of each noise type.

Another challenge is that a VSR model should be person-independent and pose-invariant. However, it is well known that deep networks rely heavily on texture~\cite{geirhos2018imagenet}. This can potentially degrade the performance, because unknown test subjects and head pose can substantially affect the appearance of the mouth. This is typically addressed by training the VSR models on a large number of subjects with varying poses. Some preliminary works on pose-invariant~\cite{cheng2020towards} and subject-independent~\cite{Wand2017} VSR have shown that this can be addressed in a more principled way, and this is another area that deserves further attention. Similarly, multi-view VSR~\cite{petridis2017end} can be beneficial, but it is not yet clear which lip views are optimal and how they should be combined. The availability of multiple cameras in meeting rooms, cars and in modern smartphones opens up a new opportunity for improving VSR systems.

The vast majority of VSR systems have focused on plain English speech. However, it is known that lip movements are affected by the context where speech is produced and the type of speech. There is evidence that lip movements tend to increase in silent speech~\cite{bicevskis2016effects} and also when speech is produced in noise (Lombard effect)~\cite{vsimko2016hyperarticulation}. Despite studies that show a performance drop when VSR models~\cite{ma19b_interspeech,petridis2018visual,heracleous2013analysis}
are tested on such conditions, this area remains unexplored. Finally, the development of non-English VSR systems that take into account the unique characteristics and accents of each language also remains an open challenge.

\subsection{Ethical Considerations}
It is important to note that VSR is a dual-use technology, which means it can have a positive impact on society as well as a negative one. Although our objective is to build VSR systems that will be beneficial for society, like the applications mentioned above, this technology can also be misused. One example is that it can be deployed for surveillance via CCTV or even via smartphone cameras, which raises privacy concerns~\cite{nyt2018,vice2021}. A potential side effect of this is that it might discourage people from speaking in public if they believe that their conversation can be intercepted by anyone carrying a camera~\cite{vice2021}. Sophisticated surveillance using VSR technology might not be possible at the moment, especially via CCTV due to the low quality of CCTV camera images, compared to the high-quality data used during training, but it should not be ignored. Cameras and VSR systems are getting better, so it might become a serious privacy concern rather soon unless automatic blurring of all faces of people who did not provide an explicit consent becomes a new standard.

Commercial applications of VSR technology are still at a very early stage. One of the very few examples is a smartphone application that aims to help speech-impaired individuals communicate and is currently being trialled in UK NHS hospitals. This is being developed by Liopa~\cite{liopa}, which also works on keyword spotting from CCTV footage. We thus argue that appropriate government regulations for VSR systems, which address privacy concerns and potential misuse, are necessary at this early stage before the technology is fully commercialized. This will allow the proper auditing of every new application before it reaches the market, so that the risks and merits can be properly communicated to users and the public. Otherwise, VSR systems may have the same fate as face recognition technology, which was commercialized without proper regulation being in place. As a consequence, a ban on using face recognition was introduced in several cities~\cite{wired,cities} and some companies either stopped offering such services or put restrictions on their use~\cite{FB,Amazon,microsoft} when the ethical concerns became widely known.

It should also be pointed out that VSR technology might be biased against specific age groups, genders, cultural backgrounds or non-native speakers. Most of the publicly available datasets have been collected from TV programmes, TED talks or YouTube videos. Hence, it is very likely that some groups are underrepresented, for example, younger people when data are collected from TV programmes or older people when data are collected from YouTube. Similarly, it is likely that people from specific cultural backgrounds or non-native speakers are also underrepresented. This will lead to VSR models that are less accurate for all these groups. Because demographic information is not available for any publicly available dataset used for training VSR models, it is not easy to verify whether such biases exist. VSR models need to be trained on demographically diverse data, including non-native speakers, to ensure similar performance across different user groups. This will lead to VSR systems whose accuracy is not lower for some users because their age, gender, cultural background or accent is underrepresented in the training data.

\section{Visual Speech Recognition}
\begin{table*}[!t]
\centering
\caption{Ablation study on the LRS2 dataset and LRS3 dataset. Models are trained on LRW+LRS2 and LRW+LRS3, respectively.}
\renewcommand\arraystretch{1.1}
\begin{tabularx}{\textwidth}{u y y}
\toprule
Method &WER on LRS2 &WER on LRS3 \\
\midrule\midrule
Our model  &29.5$\pm$0.4 &35.8$\pm$0.5 \\
\midrule
- Audio auxiliary task  &31.4$\pm$0.3 &36.6$\pm$0.3 \\
\midrule
- Visual auxiliary task &30.6$\pm$0.5 &36.9$\pm$0.5 \\
\midrule
- Audio auxiliary task, visual auxiliary task & 33.2$\pm$0.5 &37.8$\pm$0.6 \\
\midrule
- Time masking & 32.6$\pm$0.5 &38.5$\pm$0.5 \\
\midrule
- Audio auxiliary task, visual auxiliary task, time masking & 35.0$\pm$0.5 &39.1$\pm$0.4 \\
\bottomrule
\end{tabularx}
\label{tab: ablation_study_on_lrs2_and_lrs3}
\end{table*}
Our method outperforms state-of-the-art methods by a large margin for VSR in multiple languages. In what follows we explain the details of our approach and the changes that we have made to the training strategy and architecture that led to this highly improved performance.

\subsection{Datasets}
\textbf{LRS2.}~\cite{chung2017lip} describes a large-scale audio-visual English dataset collected from BBC programmes. It consists of 144,482 video clips with a total duration of 224.5 h. The videos are divided into a pre-training set with 96,318 utterances (195 h), a training set with 45,839 utterances (28 h), a validation set with 1,082 utterances (0.6 h) and a test set with 1,243 utterances (0.5 h).

\noindent\textbf{LRS3.}~\cite{afouras2018lrs3} describes the largest publicly audio-visual English dataset collected from TED talks. It contains 438.9 h with 151,819 utterances. Specifically, there are 118,516 utterances in the ‘pre-train’ set (408 h), 31,982 utterances in the ‘train-val’ set (30 h) and 1,321 utterances in the ‘test’ set (0.9 h).

\noindent\textbf{CMLR.}~\cite{zhao2019cascade} describes a large-scale audio-visual Mandarin dataset collected from a Chinese national news programme. It contains 102,072 clips with transcriptions. The training, validation and test sets contain 71,448~(60.6~h), 10,206 (8.6 h) and 20,418 (17.3 h) clips, respectively. To the best of our knowledge, CMLR is the largest publicly available dataset in Mandarin.

\noindent\textbf{CMU-MOSEAS.}~\cite{bagher-zadeh-etal-2020-cmu} describes a large-scale dataset that contains multiple languages and was collected from YouTube videos. It consists of 40,000 transcribed sentences and includes Spanish, Portuguese, German and French. We consider the Spanish videos (CM$_{\text{es}}$) with a total duration of 16.3 h. We divided the data into training and test sets, which contain 8,253 videos (15.7 h) and 329 videos (0.6 h), respectively.

\noindent\textbf{Multilingual TEDx.}~\cite{salesky21_interspeech} describes a multilingual corpus collected from TEDx talks. It covers eight languages with manual transcriptions and has a total duration of 765 h. For the purposes of this study, we consider the Spanish videos (MT$_{\text{es}}$) and use the data split proposed in~\cite{salesky21_interspeech}. We manually cleaned the dataset to exclude videos where the speaker is not visible, resulting in a total of 44,745 videos~(71.4 h) for training, 403 videos (0.7 h) for validation and 302 videos (0.5 h) for testing. It should be noted that we only use the training set in this study.

\noindent\textbf{AVSpeech.}~\cite{DBLP:journals/tog/EphratMLDWHFR18} is a large-scale audio-visual dataset consisting of 4,700 h of video in multiple languages. A pre-trained language recognition model, VoxLingua107~\cite{valk2021slt}, was first used to identify the English speaking videos. Two pre-trained ASR models, Wav2Vec2-Base-960h (\url{https://huggingface.co/facebook/wav2vec2-base-960h}) and Wav2Vec2-large-xlsr-53-english (\url{https://huggingface.co/jonatasgrosman/wav2vec2-large-xlsr-53-english}), were then used to obtain machine-generated transcriptions for these videos. We only kept the videos where the WER between the two generated transcriptions was below 60\,\%, resulting in 350,991 videos with a total duration of 641 h. The transcriptions generated by the Wav2Vec2-Base-960h model were used for these videos.

\subsection{Performance Metrics}
WER is the most common metric used in speech recognition. This measures how close the predicted word sequence is to the target word sequence. Assuming $S$ is the number of substitutions, $D$ is the number of deletions, $I$ is the number of insertions needed to get from the predicted to the target sequence and $N$ is the number of words in the target sequence, then the metric can be defined as

\begin{equation}
    WER = \frac{S+D+I}{N}
    \label{eq: error rate}
\end{equation}

Similarly to WER, we can define the CER, which measures how close the predicted and target character sequences are. In this case, $S$, $D$ and $I$ are computed at the character level and $N$ is the total number of characters.

\subsection{Pre-processing}
We used the RetinaFace~\cite{DBLP:journals/corr/abs-1905-00641} face detector and the Face Alignment Network (FAN)~\cite{bulat2017far} to detect 68 facial landmarks. The faces were then registered to a neutral reference frame using a similarity transformation to remove translation and scaling variations. A bounding box of $96 \times 96$, centred on the mouth centre, was used to crop the mouth region of interest. The cropped patch was further converted to grey-scale and normalized with respect to the overall mean and variance of the training set.

\subsection{Hyperparameter optimization}
Hyperparameter optimization aims to improve the performance of a model by fine-tuning the values of the parameters that are used to control the training process or the model architecture. Some of the most common hyperparameters that are usually optimized are the following: initial learning rate, learning rate decay parameters, number of layers, size of layers, dropout rate and the loss function weights, which are used to combine the different loss terms. Additional hyperparameters related to conformers are the number and size of the self-attention heads. We performed hyperparameter optimization on the LRS2 dataset by attempting to reduce the WER on the validation set. Our conclusion was that the parameters used in the baseline model~\cite{DBLP:journals/corr/abs-2102-06657} were already optimal, so no further improvement was observed.

The next step was to optimize other hyperparameters that might not have been exhaustively optimized, like batch-size-related parameters. Again, the parameters were chosen based on the validation set performance. Further details and results are provided in Supplementary Section~\ref{sec:SI_hyperOptim} and Supplementary Table~\ref{tab: hyperparameterOptimisation_ablationStudy_validation_set}, respectively. The results on the LRS2 and LRS3 test sets are shown in Supplementary Table~\ref{tab: hyperparameterOptimisation_ablationStudy}. Each hyperparameter was optimized independently based on the WER on the validation set of LRS2. We used the same hyperparameters for all experiments. It is clear that hyperparameter optimization results in a substantial reduction in the WER for both datasets.

\subsection{Improving LMs}
A LM determines the probability of a given sequence of characters. It is used during decoding and favours sequences that are more likely to occur. To increase the capacity of the LM we use multiple text corpora for training. We also increase the number of sequences considered during decoding (beam size is set to 40). The impact of these changes is demonstrated in Supplementary Table~\ref{tab: hyperparameterOptimisation_ablationStudy}, where the WER is reduced for both English datasets.

The score from the LM (S$_{LM}$) is incorporated in decoding as follows:
\begin{align}
S = \lambda S_{CTC} + (1-\lambda) S_{att} 
 + \beta S_{LM}
\label{eq:decode}
\end{align}
where $S_{CTC}$ and $S_{att}$ are the scores of the CTC and decoder branch, respectively, and $\lambda$ and $\beta$ correspond to the CTC and LM score weights. Additional details about the corpora used for training the LM in each language, as well as training details, are presented in Supplementary Section 5.

\subsection{Time Masking}
Data augmentation works by synthesizing additional distorted training data with the goal of reducing over-fitting. In VSR, most existing works make use of image transformations such as random cropping and horizontal flipping~\cite{DBLP:journals/corr/SimonyanZ14a, DBLP:journals/corr/abs-2102-06657, petridis2018audio}. These spatial augmentations are helpful, but they do not take into account the temporal nature of visual speech. Only a few works exist that apply temporal augmentations like deleting or duplicating frames~\cite{assael2016lipnet} or variable length augmentation~\cite{ma2020towards}.

In this Article, we propose the use of time-masking, which is commonly used in training ASR models~\cite{park2019specaugment}. It works by randomly masking $n$ consecutive frames by replacing them with the mean sequence frame. This allows the model to more effectively use contextual information and can better disambiguate similar lip movements that correspond to different phonemes. It also makes the model more robust to short missing segments. Given that there is large variance in the video lengths, especially on the LRS2 and LRS3 datasets, the number of masks used is proportional to the length of the training sequence. Specifically, we use one mask per second and, for each mask, we randomly mask up to 40\% of frames, with the masked segments chosen using a uniform distribution. Additional details about this augmentation are provided in Supplementary Section~\ref{sec:SI_TM}.

The impact of time-masking is shown in the ablation study on the LRS2 and LRS3 datasets shown in Table~\ref{tab: ablation_study_on_lrs2_and_lrs3}. Training a model without time-masking results in a substantial increase in the mean WER when compared to the full model.

\subsection{Prediction-based Auxiliary Tasks}
The standard approach to VSR relies on end-to-end training, which allows the entire model to be optimized towards the desired target. This is an attractive property and has led to impressive results, but also results in substantial challenges in training such a large model. One solution that has recently been proposed is the use of auxiliary tasks in the form of additional losses applied to intermediate layers of the model~\cite{Liu2021,toshniwal2017multitask,lee2021intermediate}. This acts as regularization, which helps the model learn better representations and leads to better generalization on test data.

Based on this observation, we propose as an auxiliary task the prediction from intermediate layers of audio and visual representations learned by pre-trained ASR and VSR models (Fig.~1c). This is inspired by the recent success of prediction tasks in self-supervised learning. In particular, good audio representations can be learned by predicting handcrafted audio features~\cite{pascual19_interspeech} or by using joint audio and visual supervision~\cite{shukla2020learning}. Similarly, visual speech representations can be learned by predicting audio features~\cite{ma21c_interspeech}. Hence, the proposed auxiliary task provides additional supervision to the intermediate layers of the model, which in turns results in better visual representations and improved performance. Mathematically, this is formulated as a regression problem where the goal is to minimize the L1 distance between the predicted and pre-trained visual and audio features. This results in the following loss term added to the loss function:

\begin{align}
    \mathcal{L}_{\textit{AUX}} &= \beta_a\left\lVert h_a(f^l(\x_v)) - g^l_a(\x_a) \right\rVert _1 \nonumber\\& + \beta_v\left\lVert h_v(f^l(\x_v)) - g^l_v(\x_v) \right\rVert _1
\label{eq: aux_loss}
\end{align}

where $\x_v$ and $\x_a$ are the visual and audio input sequences, respectively, $g_v$ and $g_a$ are the pre-trained visual and audio encoders, respectively. $f$ is the subnetwork up to layer $l$ whose intermediate representation is used as input to the audio and visual predictors $h_a$ and $h_v$, respectively. $\beta_a$ and $\beta_v$ are the coefficients for each loss term and ${\left\lVert{\cdot}\right\rVert _1}$ is the $\ell_1$-norm. 

The model performs VSR and at the same time attempts to predict audio and visual representations from intermediate layers. Hence, the final loss is simply the addition of the main VSR loss and the auxiliary loss:

\begin{align}
  \mathcal{L} = \mathcal{L}_{\textit{VSR}} + \mathcal{L}_{\textit{AUX}}
\end{align}

\begin{align}
  \mathcal{L}_{\textit{VSR}} = \alpha \mathcal{L}_{CTC} + (1-\alpha) \mathcal{L}_{att}
\label{eq:trainingCTCweight}
\end{align}

where $\mathcal{L}_{VSR}$ is the loss of the hybrid CTC/attention architecture used. $\mathcal{L}_{CTC}$ is the CTC loss, $\mathcal{L}_{att}$ the loss of the attention mechanism, and $\alpha$ controls the relative weight of each loss term. Further details about the losses are provided in Supplementary Section~\ref{sec:SI_loss}. We emphasize that the proposed method is not architecture-dependent and can also be used with other more advanced visual front ends~\cite{serdyuk2022transformer}.

The substantial impact of the auxiliary losses on performance can be observed from Table~\ref{tab: ablation_study_on_lrs2_and_lrs3}. Removing either loss, that is, either the first or second term from equation (\ref{eq: aux_loss}), leads to an increase in the mean WER for both datasets. In the case where both losses are removed, that is, no auxiliary loss is used, then the increase in the mean WER is even greater. Finally, the removal of the two losses and time-masking results in a substantial decrease in performance.

An ablation study on the effect of layer l where the auxiliary loss (equation (\ref{eq: aux_loss})) is attached is shown in Supplementary Fig.~\ref{fig:different_positions_mtl_set}. Layer 6 was found to be the optimal level based on the performance on the validation set. All results reported in all the tables are based on this configuration. Further details are presented in Supplementary Section~\ref{ssec:SI_Ablation_layer_position}.

\subsection{Using Additional Training Data}
Using larger and larger training sets with a view to reducing the WER is a recent trend in the literature. To investigate the impact of the amount of training data, we trained models on varying amounts of data. We started by training models using only the training set of each database (seventh row of Table~\ref{table: results_on_LRS2} and fourth row of Table~\ref{table: results_on_LRS3}). It is not possible to train a model from scratch on the LRS2 and LRS3 datasets, so we used curriculum learning. This means that we first used only short utterances and as training progresses we kept adding longer ones. Further details on curriculum learning are provided in Supplementary Section~\ref{sec:SI_CL}. We used a model trained for recognizing 500 English words~\cite{ma2020towards} on the LRW dataset for initialization, then fine-tuned it on the corresponding training sets of the LRS2 or LRS3 datasets (eighth row of Table~\ref{table: results_on_LRS2} and fifth row of Table~\ref{table: results_on_LRS3}). Finally, we used the models trained on LRW + LRS3 and LRW + LRS2 as initialization and fine-tuned them further on LRS2 and LRS3, respectively (ninth row of Table~\ref{table: results_on_LRS2} and sixth row of Table~\ref{table: results_on_LRS3}). It is clear that, as we use more datasets for training, the performance keeps improving. This is also the case for Spanish and Mandarin (sixth row of Table~\ref{tab: cmlr_results} and third row of Table~\ref{tab: cmumoseas_results}), even when models trained on English are used for initialization. However, the reduction in WER is smaller than in English, probably due to language mismatch.

Finally, we used a subset of the AVspeech dataset as additional training data together with the automatically generated English transcriptions. Again, the WER is reduced in all languages (tenth row of Table~\ref{table: results_on_LRS2}, seventh row of Table~\ref{table: results_on_LRS3}, last row of Table~\ref{tab: cmlr_results} and~\ref{tab: cmumoseas_results}), despite using transcriptions that contain errors, with the smallest reduction observed in Mandarin. This is not surprising, because Mandarin is much less similar to English than Spanish. These results are in line with the hypothesis that the reduction in the WER reported in recent works is mainly due to the larger datasets used for training.

\subsection{Implementation}
Our experiments were implemented using an open-source toolkit, ESPNet~\cite{watanabe2018espnet}. We trained the models with the Adam optimizer~\cite{kingma2014adam} with $\beta_{1}=0.9$, $\beta_{2}=0.98$ and $\epsilon=10^{-9}$. The learning rate increases linearly in the first 25,000 steps, yielding a peak learning rate of 0.0004 and thereafter decreasing in proportional to the inverse square root of the step number. The network was trained for 50 epochs with a batch size of 16. We used the model averaged over the last ten checkpoints for evaluation. Details regarding the network architecture are provided in Supplementary Section~\ref{sec:SI_arch}.

\section{Conclusions}
In this Article we have presented our approach for VSR and demonstrated that state-of-the-art performance can be achieved not only by using larger datasets, which is the current trend in the literature, but also by carefully designing a model. We have highlighted the importance of hyperparameter optimization, which can further improve the performance of existing architectures. We have then shown the importance of time-masking, which forces the network to focus more in the context. We have also proposed a new architecture based on auxiliary tasks where the VSR model also predicts audio-visual representations learned by pre-trained ASR and VSR models. Finally, we have provided evidence that using larger datasets improves the performance, which is in line with recent works in this field. Our approach outperforms all existing VSR works trained on publicly available datasets in English, Spanish and Mandarin, by a large margin.

\section*{Data Availability} 

The datasets used in the current study are available from the original authors on the LRS2 (\url{https://www.robots.ox.ac.uk/~vgg/data/lip_reading/lrs2.html}), LRS3 (\url{https://www.robots.ox.ac.uk/~vgg/data/lip_reading/lrs3.html}), CMLR (\url{https://www.vipazoo.cn/CMLR.html}), Multilingual  (\url{http://www.openslr.org/100}), and CMU-MOSEAS (\url{http://immortal.multicomp.cs.cmu.edu/cache/multilingual}) repositories. Qualitative results and the list of cleaned videos for the training and test sets of CMU-MOSEAS and Multilingual TEDx are available on the authors’ GitHub repository (\url{https://mpc001.github.io/lipreader.html}).

\section*{Code Availability} 
Pre-trained networks and testing code are available on a GitHub repository (\url{https://mpc001.github.io/lipreader.html}) or at Zenodo~\cite{pingchuan_ma_2022_6651667} under an Attribution-NonCommercial-NoDerivatives 4.0 International (CC BY-NC-ND 4.0) licence.

\section*{Acknowledgements}
All training, testing and ablation studies were conducted at Imperial College London.

\section*{Authors' Contributions} 
The code was written by P.M., and the experiments were conducted by P.M. and S.P. The manuscript was written by P.M., S.P. and M.P. M.P. supervised the entire project.

\section*{Competing Interests}
The authors declare no competing interests.

\section*{Additional Information} 
Correspondence and requests for materials should be addressed to Pingchuan Ma.

\bibliographystyle{naturemag}
\bibliography{references}

\clearpage

\setcounter{section}{-1}\stepcounter{section}
\setcounter{table}{-1}\stepcounter{table}
\setcounter{equation}{-1}\stepcounter{equation}
\setcounter{figure}{-1}\stepcounter{figure}
\renewcommand{\thesection}{S\arabic{section}}
\renewcommand{\thetable}{S\arabic{table}}
\renewcommand\theequation{S\arabic{equation}}
\renewcommand\thefigure{S\arabic{figure}}

\section{Datasets Details}
Details about the audio-visual datasets used in this study are presented in Supplementary Table~\ref{tab: SI_datasets}. It is clear that the non-publicly available datasets are one to two orders of magnitude larger than the publicly available ones. 

\section{Architecture Details}
\label{sec:SI_arch}
The model consists of 4 modules, a front-end encoder, VSR encoder in Fig.~1c
, a back-end encoder, a hybrid CTC and transformer decoder and two predictors. In particular, the encoder receives as input the raw images and maps them to visual speech representations which are fed to the back-end encoder. This is followed by a CTC and transformer decoder which generates the predicted characters. Finally, the features extracted from the middle position of the back-end encoder flow through two separate predictors to predict visual and acoustic speech representations from pre-trained VSR and ASR models, respectively.

The \textbf{front-end encoder} consists of a 3D convolutional layer with a kernel size of $5 \times 7 \times 7$ followed by a  ResNet-$18$~\cite{he2016deep, stafylakis2017combining}.  Let $B\times T\times H\times W$ be the input tensor to the visual front-end module, where $B$, $T$, $H$, and $W$ correspond to batch size, number of frames, height and width, respectively. The visual features at the top of the residual blocks are aggregated along the spatial dimension by a global average pooling layer, resulting in a feature output of dimensions $B\times C\times T$, where $C$ indicates the channel dimensionality. The Swish activation functions is used in all layers. The detailed architecture can be seen in Supplementary Table~\ref{net:vsr_frontend}.

\begin{table}[!t]
\centering
\caption{Details of Audio-Visual Datasets used in this work. CM$_{\text{xx}}$ and MT$_{\text{xx}}$ denote the particular language parts of the CMU-MOSEAS and Multilingual TEDx datasets, respectively, where $\text{xx}$ denotes the standard language codes, conforming to the ISO 639-1 standard.}
\renewcommand\arraystretch{1.0}
\begin{tabularx}{\columnwidth}{j k k}
\toprule
Dataset &Transcription  &Hours  \\
\midrule\midrule
\multicolumn{3}{c}{\textit{Publicly Available Datasets}} \\ \midrule
LRW~\cite{chung2016lip} &\cmark &157  \\
\midrule
LRS2~\cite{chung2017lip} &\cmark & 223 \\
\midrule
LRS3~\cite{afouras2018lrs3} &\cmark & 438 \\
\midrule
CMLR~\cite{zhao2019cascade} &\cmark & 61 \\
\midrule
MT$_{\text{es}}$~\cite{salesky21_interspeech} &\cmark & 71 \\
\midrule
MT$_{\text{it}}$\cite{salesky21_interspeech} &\cmark &46 \\
\midrule
MT$_{\text{pt}}$~\cite{salesky21_interspeech} &\cmark &81 \\
\midrule
MT$_{\text{fr}}$~\cite{salesky21_interspeech} &\cmark &85 \\
\midrule
CM$_{\text{es}}$~\cite{bagher-zadeh-etal-2020-cmu} &\cmark & 16 \\
\midrule
CM$_{\text{pt}}$~\cite{bagher-zadeh-etal-2020-cmu} &\cmark & 18 \\
\midrule
CM$_{\text{fr}}$~\cite{bagher-zadeh-etal-2020-cmu} &\cmark & 15 \\
\midrule
AVSpeech~\cite{DBLP:journals/tog/EphratMLDWHFR18} &\xmark & 641 \\
\midrule
\multicolumn{3}{c}{\textit{Non-Publicly Available Datasets}} \\
\midrule
MVLRS~\cite{chung2017lip} &\cmark &730 \\
\midrule
LSVSR~\cite{shillingford2019large} &\cmark &3\,886 \\
\midrule
YT-31k~\cite{makino2019recurrent} &\cmark &31\,000 \\
\midrule
YT-90k~\cite{serdyuk2021audiovisual} &\cmark &90\,000 \\
\midrule
VoxCeleb2$^{\text{clean}}$~\cite{afouras2020asr} &\xmark &334 \\
\bottomrule
\end{tabularx}
\label{tab: SI_datasets}
\end{table}
\begin{table*}[!tb]
\caption{The architecture of the front-end encoder of the VSR model. The filter shapes are denoted by $\{ \text{Temporal Size}\times\text{Spatial Size}^2, \text{Channels} \}$ and $\{ \text{Spatial\ Size}^2, \text{Channels} \}$ for 3D convolutional and 2D convolutional Layers , respectively. The sizes correspond to [Batch Size, Channels, Sequence Length, Height, Width] and [Batch Size $\times$ Sequence Length, Channels, Height, Width], for 3D and 2D convolutional layers, respectively. $T_v$ denotes the number of input frames.}
\begin{center}{\scalebox{1.0}{

\renewcommand{\arraystretch}{1.62}
\begin{tabular}{c|p{5.5cm}<{\centering}|p{3cm}<{\centering}|p{3cm}<{\centering}}
\hlineB{2}
Component Name
&Layer Type &Input Size &Output Size \\
\hline

\multirow{ 2}{*}{$\text{Stem}_1$}
&
Conv 3D, 5 $\times$ 7$^2$, 64 &[B, 1, T$_v$, 88, 88] &[B, 64, T$_v$, 44, 44]    \\
\cline{2-4}
&
3D Max Pooling, 1 $\times$ 3$^2$ &[B, 64, T$_v$, 44, 44] &[B, 64, T$_v$, 22, 22]   \\
\cline{1-4}
$\text{Reshape}$&- &[B, 64, T$_v$, 22, 22] &[B$\times$T$_v$, 64, 22, 22] 
\\
\cline{1-4}
$\text{Residual Block}_2$
&
{$\begin{bmatrix} \text{Conv 2D}, 3^2, 64 \\ \text{Conv 2D}, 3^2, 64 \\ \end{bmatrix} \times 2$}
&[B$\times$T$_v$, 64, 22, 22] &[B$\times$T$_v$, 64, 22, 22]
\\
\cline{1-4}

$\text{Residual Block}_3$
& {$\begin{bmatrix} \text{Conv 2D}, 3^2, 128 \\ \text{Conv 2D}, 3^2, 128 \\ \end{bmatrix} \times 2$}
&[B$\times$T$_v$, 64, 22, 22] &[B$\times$T$_v$, 128, 11, 11]\\
\cline{1-4}

$\text{Residual Block}_4$
& {$\begin{bmatrix} \text{Conv 2D}, 3^2, 256 \\ \text{Conv 2D}, 3^2, 256 \\ \end{bmatrix} \times 2$}
&[B$\times$T$_v$, 128, 11, 11] &[B$\times$T$_v$, 256, 6, 6]\\
\cline{1-4}

$\text{Residual Block}_5$
& {$\begin{bmatrix} \text{Conv 2D}, 3^2, 512 \\ \text{Conv 2D}, 3^2, 512 \\ \end{bmatrix} \times 2$}
&[B$\times$T$_v$, 256, 6, 6] &[B$\times$T$_v$, 512, 3, 3]\\
\cline{1-4}
Aggregation&
$\text{2D Global Average Pooling}$
&[B$\times$T$_v$, 512, 3, 3] &[B$\times$T$_v$, 512, 1, 1]\\
\cline{1-4}
$\text{Reshape}$ &-
&[B$\times$T$_v$, 512, 1, 1] &[B, 512, T$_v$]\\

\hlineB{2}
\end{tabular}}}
\end{center}
\label{net:vsr_frontend}

\end{table*}
\begin{table*}[!tb]
\caption{The architecture of the front-end encoder of the ASR model. The filter shapes are denoted by $\{ \text{Temporal Size}, \text{Channels} \}$ for 1D Convolutional Layers, respectively. The sizes correspond to [Batch Size, Channels, Sequence Length]. $T_a$ denotes the length of audio waveforms.}
\begin{center}{\scalebox{1.0}{
\renewcommand{\arraystretch}{1.62}
\begin{tabular}{c|p{5.5cm}<{\centering}|p{3cm}<{\centering}|p{3cm}<{\centering}}
\hlineB{2}
Component Name
&Layer Type &Input Size &Output Size \\
\hline

\text{Stem}$_1$
&
Conv 1D, 80, 64 &[B, 1, T$_a$] &[B, 64, T$_a//4$]    \\
\cline{1-4}
$\text{Residual Block}_2$
&
{$\begin{bmatrix} \text{Conv 1D}, 3, 64 \\ \text{Conv 1D}, 3, 64 \\ \end{bmatrix} \times 2$} 
&[B, 64, T$_a//4$] &[B, 64, T$_a//4$]
\\
\cline{1-4}

$\text{Residual Block}_3$
&{$\begin{bmatrix} \text{Conv 1D}, 3, 128 \\ \text{Conv 1D}, 3, 128 \\ \end{bmatrix} \times 2$} 
&[B, 64, T$_a//4$] &[B, 128, T$_a//8$]\\
\cline{1-4}

$\text{Residual Block}_4$
&{$\begin{bmatrix} \text{Conv 1D}, 3, 256 \\ \text{Conv 1D}, 3, 256 \\ \end{bmatrix} \times 2$} 
&[B, 128, T$_a//8$] &[B, 256, T$_a//16$]\\
\cline{1-4}

$\text{Residual Block}_5$
&{$\begin{bmatrix} \text{Conv 1D}, 3, 512 \\ \text{Conv 1D}, 3, 512 \\ \end{bmatrix} \times 2$} 
&[B, 256, T$_a//16$] &[B, 512, T$_a//32$]\\
\cline{1-4}
Aggregation
&1D Average Pooling, Stride 20
&[B, 512, T$_a//32$] &[B, 512, T$_a//640$]\\
\hlineB{2}
\end{tabular}}}
\end{center}
\label{net:asr_frontend}
\end{table*}

The \textbf{back-end encoder} starts with a positional embedding module, followed by a stack of 12 conformer blocks. The positional embedding module is a linear layer, which projects the features from the output of ResNet-18 to a 256-dimensional space. The transformed features are further injected with relative position information~\cite{dai2019transformer}. In each conformer block, a feed-forward module, a self-attention module, a convolution module, and a second feed-forward module are stacked in order. Specifically, the feed-forward module is comprised of a linear layer, which projects the features to a higher 2048-dimensional space, followed by a Rectified Linear Unit (ReLU) activation function, a dropout layer with a probability of 0.1, and a second linear layer with output dimension of 256. Half-step residual connections are also used in each feed-forward module. The self-attention module is capable of modeling global dependencies among elements. The module maps the query and a set of key-value pairs through an attention map, which focuses on different parts of the input. Instead of performing a single attention function, a multi-head mechanism is leveraged with different linear projections to a lower $64$-dimensional space. The attention function is performed in parallel on each head and the outputs are concatenated into a 256-dimensional space and once again projected into the final values. The convolutional module, which excels at capturing local patterns efficiently, is composed of an 1D point-wise convolutional layer, Gated Linear Units (GLU)~\cite{dauphin2017language}, an 1D depth-wise convolutional layer, a batch normalisation layer, a swish activation layer, a 1D point-wise convolutional layer, and a layer normalisation layer. The combination of self-attention and convolution is capable of better capturing both local and global temporal information compared to the standard transformer architecture~\cite{gulati2020conformer}.

The \textbf{decoder} is composed of an embedding module and a set of residual multi-head attention blocks. It takes as input the encoded sequence and the prefixes of the target sequence. First, the prefixes from index 1 to $l$ - 1 are projected to embedding vectors, where $l$ is the target length index. The absolute positional encoding~\cite{vaswani2017attention} is also added to the embedding. Next, the embedding is fed to a stack of multi-head attention blocks. Each block consists of a self-attention module, an encoder-decoder attention module and a feed-forward module. Layer normalisation is added before each module. Specifically, the self-attention module is slightly different from the one in the encoder where future positions at its attention matrix are masked out, followed by an encoder-decoder attention, which helps the decoder to focus on the relevant part of the input. This attention receives the features from the previous self-attention module as $Q$ and the features from the encoder as $K$ and $V$ ($K=V$). The features are further fed to a feed-forward module, which is the same as the one used in the encoder. Finally, a layer normalisation and a linear layer are added which predict the posterior distribution of the next generated token.

A \textbf{linear layer} with a softmax function, which maps the encoded features to the predicted character sequence is also used on top of the back-end encoder . This layer is trained with the CTC loss.

The \textbf{predictor} is a linear layer which takes as input the features at the middle block ($6$th) of the back-end encoder and predicts the corresponding audio/visual features from the pre-trained ASR/VSR models. Separate predictors are employed for each prediction task. Both the input and output dimensions of the linear layer are $256$.

\section{Pre-trained VSR and ASR models}
\label{sec:SI_VSR_ASR_Models}
The pre-trained ASR and VSR models are shown in Fig.~\ref{fig:av_architecture}a and ~\ref{fig:av_architecture}b, respectively. The pre-trained VSR model has exactly the same architecture as the full model described in Supplementary Section~\ref{sec:SI_arch} but does not include any predictors. The pre-trained ASR model replaces the VSR encoder with an ASR encoder and its architecture can be seen in Fig.~\ref{fig:av_architecture}d and Supplementary Table~\ref{net:asr_frontend}. It should be noted that these models are always trained on the same data as the full model. Then the pre-trained ASR/VSR encoders and some conformer layers are frozen and their internal representations are used as targets for the audio and visual predictors as shown in Fig.~\ref{fig:av_architecture}c. The performance of the pre-trained models for all languages can be seen in Supplementary Tables~\ref{table: supplemental_results_on_LRS2},~\ref{table: supplemental_results_on_LRS3},~\ref{tab: supplemental_cmlr_results} and~\ref{tab: supplemental_cmumoseas_results}.

\begin{table*}[!t]
\caption{Performance (Mean$\pm$Std.) of the pre-trained ASR and VSR models on the LRS2 dataset.}
\renewcommand\arraystretch{1.0}
\begin{tabularx}{2\columnwidth}{f q y y y }
\toprule
Method &Training Sets & Full Model  &Pre-trained VSR model &Pre-trained ASR model \\
\midrule\midrule
Ours &LRS2 &\bf 33.6$\pm$0.5 &33.4$\pm$0.3 &4.0$\pm$0.4  \\
\midrule
Ours &LRW+LRS2  &\bf 29.5$\pm$0.4 &33.2$\pm$0.5 &3.9$\pm$0.2   \\
\midrule
Ours &LRW+LRS2+LRS3 &\bf 27.6$\pm$0.2  &29.3$\pm$0.4 &3.7$\pm$0.1   \\
\midrule
Ours &LRW+LRS2+LRS3+AVSpeech &\bf 25.8$\pm$0.4 &29.3$\pm$0.4 &3.7$\pm$0.1 \\
\bottomrule
\end{tabularx}
\label{table: supplemental_results_on_LRS2}
\end{table*}

\begin{table*}[!t]
\renewcommand\arraystretch{1.0}
\caption{Performance (Mean$\pm$Std.) of the pre-trained ASR and VSR models on the LRS3 dataset.}
\begin{tabularx}{2\columnwidth}{f q y y y }
\toprule
Method &Training Sets & Full Model  &Pre-trained VSR model  &Pre-trained ASR model \\
\midrule\midrule
Ours &LRS3 &\bf 38.6$\pm$0.4 &38.7$\pm$0.5 &2.3$\pm$0.1  \\
\midrule
Ours &LRW+LRS3&\bf 35.8$\pm$0.5 &37.8$\pm$0.6 &2.2$\pm$0.1 \\
\midrule
Ours &LRW+LRS2+LRS3&\bf 34.9$\pm$0.2 &35.2$\pm$0.2 &2.0$\pm$0.2 \\
\midrule
Ours &LRW+LRS2+LRS3+AVSpeech &\bf 32.1$\pm$0.3 &35.2$\pm$0.2 &2.0$\pm$0.2 \\
\midrule
\end{tabularx}
\label{table: supplemental_results_on_LRS3}
\end{table*}
\begin{table*}[!t]
\centering
\caption{Performance (Mean$\pm$Std.) of the pre-trained ASR and VSR models on the CMLR dataset.}
\renewcommand\arraystretch{1.0}
\begin{tabularx}{2\columnwidth}{f q y y y }
\toprule
Method &Training Sets & Full Model  &Pre-trained VSR model  &Pre-trained ASR model \\
\midrule
Ours &CMLR &\bf 9.1$\pm$0.05 &10.7$\pm$0.06 & 2.5$\pm$0.03  \\
\midrule
Ours &LRW+LRS2+LRS3+CMLR &\bf 8.2$\pm$0.06 & 9.0$\pm$0.05 & 2.2$\pm$0.03 \\
\midrule
Ours &LRW+LRS2+LRS3+AVSpeech+CMLR &\bf 8.1$\pm$0.05 &8.9$\pm$0.08 & 2.2$\pm$0.03 \\
\bottomrule
\end{tabularx}
\label{tab: supplemental_cmlr_results}
\end{table*}
\begin{table*}[!t]
\centering
\caption{Performance (Mean$\pm$Std.) of the pre-trained ASR and VSR models on the CMU-MOSEAS-Spanish (CM$_{\text{es}}$) dataset.
}
\renewcommand\arraystretch{1.0}
\begin{tabularx}{2\columnwidth}{f q y y y }
\toprule
Method &Training Sets & Full Model  &Pre-trained VSR model  &Pre-trained ASR model \\
\midrule
Ours &LRW+CM$_{\text{es}}$+MT$_{\text{es}}$ &\bf 51.5$\pm$0.8 &53.2$\pm$0.4 &16.3$\pm$0.3 \\
\midrule
Ours &LRW+LRS2+LRS3+CM$_{\text{es}}$+MT$_{\text{es}}$ &\bf 47.4$\pm$0.2 &47.5$\pm$0.6 &15.4$\pm$0.1 \\ \midrule
Ours &LRW+LRS2+LRS3+AVSpeech+CM$_{\text{es}}$+MT$_{\text{es}}$ &\bf 44.6$\pm$0.6 & 45.3$\pm$0.4 & 15.4$\pm$0.1 \\
\bottomrule
\end{tabularx}
\label{tab: supplemental_cmumoseas_results}
\end{table*}
\begin{table}[!t]
\centering
\caption{Investigation of the impact of hyperparameters and Language Model (LM) choices on the validation set of the LRS2 dataset.}
\renewcommand\arraystretch{1.2}
\begin{tabularx}{\columnwidth}{u k}
\toprule
Method &WER \\
\midrule\midrule
CM-seq2seq~\cite{DBLP:journals/corr/abs-2102-06657} - Baseline & 47.7$\pm$0.5  \\
\midrule
\enskip\enskip + Hyperparameter Optimisation & 45.6$\pm$0.4 \\
\midrule
\enskip\enskip\enskip\enskip + Improved LM & 44.1$\pm$0.5  \\
\bottomrule
\end{tabularx}
\label{tab: hyperparameterOptimisation_ablationStudy_validation_set}
\end{table}
\begin{table*}[!t]
\centering
\caption{Investigation of the impact of hyperparameters and  Language Model (LM) choices on the LRS2 dataset and LRS3 dataset.}
\renewcommand\arraystretch{1.1}
\begin{tabularx}{\textwidth}{u y y}
\toprule
Method &WER on LRS2 &WER on LRS3 \\
\midrule\midrule
CM-seq2seq~\cite{DBLP:journals/corr/abs-2102-06657} - Baseline & 37.8$\pm$0.5 & 44.9$\pm$0.8 \\
\midrule
\enskip\enskip + Hyperparameter Optimisation & 35.9$\pm$0.5 &40.6$\pm$0.8 \\
\midrule
\enskip\enskip\enskip\enskip + Improved LM & 35.0$\pm$0.5 &39.1$\pm$0.4  \\
\bottomrule
\end{tabularx}
\label{tab: hyperparameterOptimisation_ablationStudy}
\end{table*}

\section{Hyperparameter Optimization}
\label{sec:SI_hyperOptim}
The main hyper-parameter that was found to have a significant impact on performance was the batch size. We observed that increasing the batch size from 8 to 16 led to reduced WER on the validation set of the LRS2 dataset (see Supplementary Table~\ref{tab: hyperparameterOptimisation_ablationStudy_validation_set}). The same pattern is also observed on the LRS2 and LRS3 test sets (see Supplementary Table~\ref{tab: hyperparameterOptimisation_ablationStudy}). There is also one more hyper-parameter which controls the batch size based on the length of the sequences. In other words, if some sequences are too long then the batch is halved. We found that increasing this threshold from 150  to 220 frames also improved the performance. We could not increase these two hyper-parameters even further due to GPU memory constraints but it is likely that the WER will be reduced even more.

\section{Language Models}
\label{sec:SI_LM}
We train six monolingual transformer-based language model~\cite{irie2019language} for 50 epochs. The English language model is trained by combining the training sets of LibriSpeech ($960$\,h)~\cite{panayotov2015librispeech}, pre-training and training sets of LRS2~\cite{chung2017lip} and LRS3~\cite{afouras2018lrs3}, TED-LIUM 3~\cite{DBLP:conf/specom/HernandezNGTE18}, Voxforge (English) and Common Voice (English)~\cite{DBLP:conf/lrec/ArdilaBDKMHMSTW20}, with a total of 166 million characters. The Mandarin language model is trained by combining the CMLR~\cite{zhao2019cascade} and news2016zh, with a total of 153 million characters. The Spanish language model is trained by combining the Spanish corpus from Multilingual TEDx~\cite{salesky21_interspeech}, Common Voice~\cite{DBLP:conf/lrec/ArdilaBDKMHMSTW20} and Multilingual LibriSpeech~\cite{Pratap2020MLSAL}, with a total of 192 million characters. The Italian language model is trained by combining the Italian corpus from Multilingual TEDx~\cite{salesky21_interspeech}, Common Voice~\cite{DBLP:conf/lrec/ArdilaBDKMHMSTW20} and Multilingual LibriSpeech~\cite{Pratap2020MLSAL}, with a total of 252 million characters. The Portuguese language model is trained by combining the Portuguese corpus from Multilingual TEDx~\cite{salesky21_interspeech}, Common Voice~\cite{DBLP:conf/lrec/ArdilaBDKMHMSTW20} and Multilingual LibriSpeech~\cite{Pratap2020MLSAL}, with a total of 85 million characters. The French language model is trained by combining the French corpus from Multilingual TEDx~\cite{salesky21_interspeech}, Common Voice~\cite{DBLP:conf/lrec/ArdilaBDKMHMSTW20} and Multilingual LibriSpeech~\cite{Pratap2020MLSAL}, with a total of 945 million characters. In our work, we set $\lambda$ and $\beta$ from equation (\ref{eq: aux_loss}) to 0.1 and \{English: 0.6, Mandarin: 0.3, Spanish: 0.4, Italian: 0.5, Portuguese: 0.3, French: 0.3\}, respectively. The impact of the improved English language model on the validation set of the LRS2 dataset can be seen in Supplementary Table~\ref{tab: hyperparameterOptimisation_ablationStudy_validation_set}. Results on the LRS2 and LRS3 test sets can be seen in Supplementary Table~\ref{tab: hyperparameterOptimisation_ablationStudy}.

\section{Time Masking}
\label{sec:SI_TM}
We mask $n$ consecutive frames with the mean frame of the video. The duration $t_n$ is chosen from 0 to an upper bound $n_{\text{max}}$ using a uniform distribution. Since there is a large variance in the video lengths of the LRS2 and LRS3 datasets, we set the number of masks proportional to the sequence length. Specifically, we use one mask per second, and for each mask, the maximum duration $n_{\text{max}}$ is set to 0.4 seconds.

\section{Loss Functions}
\label{sec:SI_loss}
To map input sequences $\x=[x_1, ..., x_T]$ such as audio or visual streams to corresponding target characters $\y=[y_1, ..., y_L]$, we consider a hybrid CTC/attention architecture~\cite{watanabe2017hybrid} in this paper, where $T$, $L$ are the lengths of the input sequence and target character sequence, respectively. The CTC loss assumes conditional independence between the output predictions and the estimated sequence posterior has the form of $P_{CTC}(\y|\x) \approx \prod_{t=1}^Tp(y_t|\x)$. The CTC loss from equation (\ref{eq:trainingCTCweight}) is defined as follows:

\begin{align}
\mathcal{L}_{\textit{CTC}} = - \text{log} P_{CTC}(\y|\x) 
\end{align}
An attention-based encoder-decoder model gets rid of this assumption by directly estimating the posterior on the basis of the chain rule and has a form of $P_{att}(\y|\x) \approx \prod_{l=1}^Lp(y_l|y_{<l}, \x)$.

In this case the $\mathcal{L}_{\textit{att}}$ from equation is:
\begin{align}
\mathcal{L}_{\textit{att}} = - \text{log} P_{att}(\y|\x)
\end{align}

The objective function of speech recognition is performed by a linear combination of the CTC loss and a cross-entropy loss as shown in equation (\ref{eq:trainingCTCweight}). The $\alpha$ value used in this work is 0.1.

A grid search was performed for the parameters $\beta_a$ and~$\beta_v$ used in the auxiliary loss (equation (\ref{eq: aux_loss})). The values that resulted in the best performance in the validation set of the LRS2 dataset are the following: $\beta_a$ = 0.4 and~$\beta_v$~=~0.4. These values are used for all experiments.

\begin{figure}[!t]
    \centering
    \includegraphics[width=\columnwidth, trim=3mm 5mm 2mm 2mm, clip]{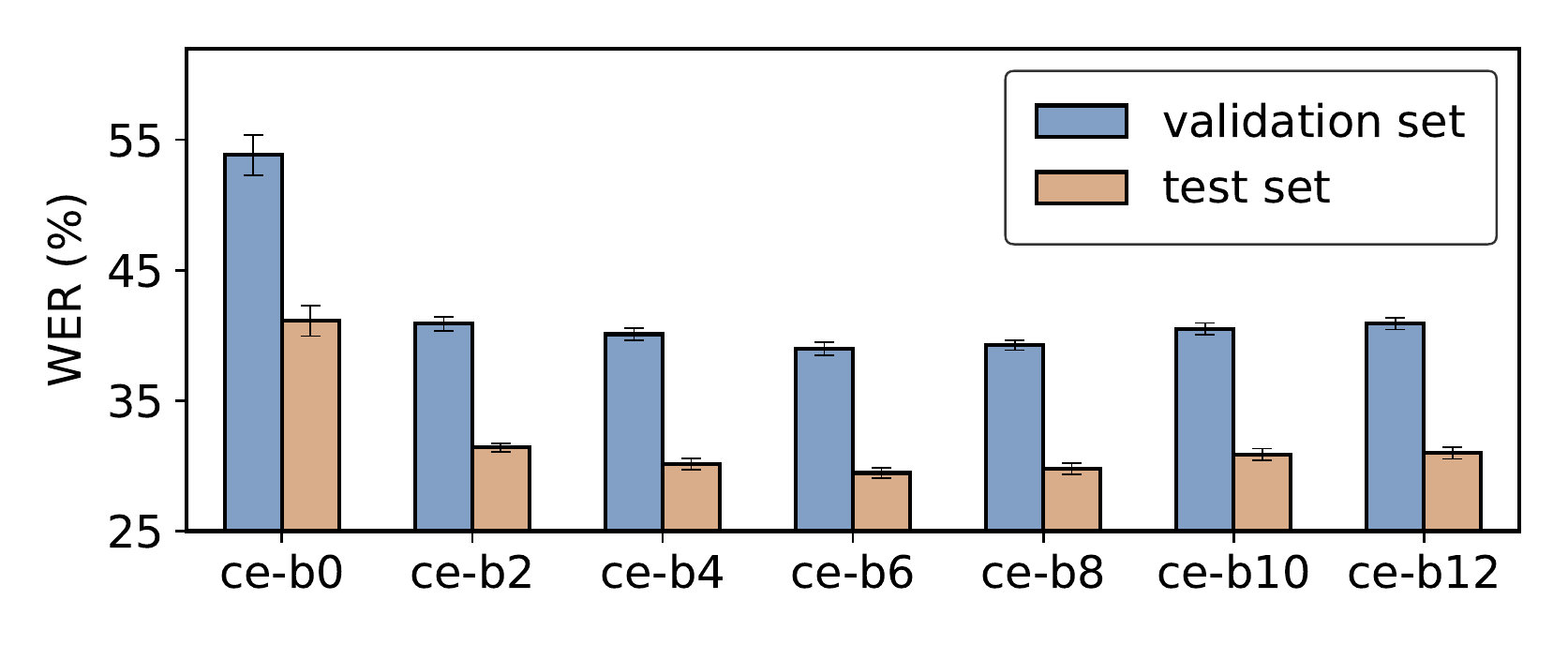}
    \caption{Performance of visual speech recognition on both the validation set and test set of LRS2 as a function of the layer where the auxiliary loss is attached (see equation \ref{eq: aux_loss}). ``ce-b0'' to ``ce-b12'' refer to the conformer layers from bottom to top.}
    \label{fig:different_positions_mtl_set}
\end{figure}
\begin{table}[!t]
\centering
\caption{Results of curriculum learning experiments on the LRS2 dataset.}
\renewcommand\arraystretch{1.1}
\begin{tabularx}{\columnwidth}{k k k}
\toprule
Video Length in Frames &WER on the Validation Set & WER on the Test Set \\
\midrule\midrule
\multicolumn{3}{c}{\textit{Baseline VSR Model}}\\
\midrule
0-100 & 65.1$\pm$0.2 &52.7$\pm$0.8 \\
\midrule
0-150 & 54.0$\pm$0.7 &44.2$\pm$0.5 \\
\midrule
0-300 & 46.0$\pm$0.6 &36.3$\pm$0.4 \\
\midrule
0-450 & 43.6$\pm$0.5 &34.3$\pm$0.5 \\
\midrule
0-600 & 42.4$\pm$0.4 &33.7$\pm$0.4 \\
\midrule
\multicolumn{3}{c}{\textit{VSR Model with Auxiliary Workers}}\\
\midrule
0-100 & 51.9$\pm$0.3 &41.5$\pm$0.5 \\
\midrule
0-150 & 46.2$\pm$0.4 &36.1$\pm$0.3 \\
\midrule
0-300 & 43.3$\pm$0.2 &34.4$\pm$0.2 \\
\midrule
0-450 & 42.6$\pm$0.3 &34.6$\pm$0.5 \\
\midrule
0-600 & 42.0$\pm$0.3 &33.4$\pm$0.3 \\
\bottomrule
\end{tabularx}
\label{tab: curriculum_learning_lrs2}
\end{table}
\begin{table}[!t]
\centering
\caption{Results of curriculum learning experiments on the LRS3 dataset.}
\renewcommand\arraystretch{1.1}
\begin{tabularx}{\columnwidth}{s s}
\toprule
Video Length in Frames &WER on the Test Set \\
\midrule\midrule
\multicolumn{2}{c}{\textit{Baseline VSR model}}\\
\midrule
0-100 & 75.2$\pm$0.4 \\
\midrule
0-150 & 53.3$\pm$0.7 \\
\midrule
0-300 & 43.0$\pm$0.4 \\
\midrule
0-450 & 39.9$\pm$0.6 \\
\midrule
0-600 & 38.7$\pm$0.5 \\
\midrule
\multicolumn{2}{c}{\textit{VSR Model with Auxiliary Workers}}\\
\midrule
0-100 & 57.7$\pm$0.4 \\
\midrule
0-150 & 46.8$\pm$0.1 \\
\midrule
0-300 & 40.8$\pm$0.6 \\
\midrule
0-450 & 39.7$\pm$0.4 \\
\midrule
0-600 & 38.6$\pm$0.4 \\
\bottomrule
\end{tabularx}
\label{tab: curriculum_learning_lrs3}
\end{table}

\section{Curriculum Learning}
\label{sec:SI_CL}
The end-to-end model was trained from scratch, resulting in poor performance on LRS2 and LRS3. This is likely due to the vast amount of very long utterances featured in LRS2 and LRS3, which makes learning from scratch especially challenging. We have found that the issue can be resolved by progressively training the end-to-end model, starting with short utterances and  then using longer ones during training. This approach is commonly called curriculum learning (CL). In this paper, the model is initially trained with a subset of labelled training data, consisting of videos shorter than 100 frames. Then this model is used for initialisation when using utterances with up to 150 frames for training. This process is repeated for 3 more rounds where the length of training sequences is 300, 450, and 600 frames, respectively.

Results for each round of curriculum learning can be seen in Supplementary Tables~\ref{tab: curriculum_learning_lrs2} and~\ref{tab: curriculum_learning_lrs3}.

\begin{table*}[!t]
\centering
\caption{Results on the Multilingual TEDx-Spanish (MT$_{\text{es}}$) dataset. `Mean$\pm$Std.' refers to the mean WER over ten runs and the corresponding standard deviation, while `Best' denotes the best (lowest) WER.}
\renewcommand\arraystretch{1.4}
\begin{tabularx}{2\columnwidth}{j v k o y y}
\toprule
Method &Pre-training Set &Training Set &Training Sets Total Size (hours) &Mean$\pm$Std. &Best \\
\midrule
CM-seq2seq~\cite{DBLP:journals/corr/abs-2102-06657} &LRW &CM$_{\text{es}}$+MT$_{\text{es}}$ &244 & 66.4$\pm$0.8 & 65.2 \\
\midrule
Ours &LRW &CM$_{\text{es}}$+MT$_{\text{es}}$ &244 &\bf 60.8$\pm$0.8 &\bf 60.3 \\
\midrule
Ours &LRW+LRS2+LRS3 &CM$_{\text{es}}$+MT$_{\text{es}}$ &905 &\bf 56.9$\pm$0.5 &\bf 56.5 \\
\midrule
Ours &LRW+LRS2+LRS3+AVSpeech &CM$_{\text{es}}$+MT$_{\text{es}}$ &1\,546 &\bf 56.6$\pm$0.3 &\bf 56.3 \\
\bottomrule
\end{tabularx}
\label{tab: supplemental_MTspanish_results}
\end{table*}
\begin{table*}[!t]
\centering
\caption{Results on the Multilingual TEDx-Italian (MT$_{\text{it}}$) dataset. `Mean$\pm$Std.' refers to the mean WER over ten runs and the corresponding standard deviation, while `Best' denotes the best (lowest) WER.}
\renewcommand\arraystretch{1.4}
\begin{tabularx}{2\columnwidth}{j v k o y y}
\toprule
Method &Pre-training Set &Training Set &Training Sets Total Size (hours) &Mean$\pm$Std. &Best \\
\midrule
CM-seq2seq~\cite{DBLP:journals/corr/abs-2102-06657} &LRW &MT$_{\text{it}}$ &203 & 71.5$\pm$0.4 & 70.9 \\
\midrule
Ours &LRW &MT$_{\text{it}}$ &203 &\bf 65.9$\pm$0.5 &\bf 65.2 \\
\midrule
Ours &LRW+LRS2+LRS3 &MT$_{\text{it}}$ &864 &\bf 58.7$\pm$0.3 &\bf 58.2 \\
\midrule
Ours &LRW+LRS2+LRS3+AVSpeech &MT$_{\text{it}}$ &1\,505 &\bf 57.9$\pm$0.7 &\bf 57.4 \\
\bottomrule
\end{tabularx}
\label{tab: supplemental_MTitalian_results}
\end{table*}

\section{Additional Results}
\label{sec:SI_Addtional_Results}
\subsection{Ablation Study on the Effect of Layer Position}
\label{ssec:SI_Ablation_layer_position}
We investigate the effect of the layer $l$ where the auxiliary loss (equation (\ref{eq: aux_loss})) is attached. The position of layer varies from 0 to 12 at intervals of 2. Layer 6 was found to be the optimal level on the validation set of LRS2. Results are presented in Supplementary Fig.~\ref{fig:different_positions_mtl_set}.

\subsection{Results on Spanish}
Results on the Multilingual TEDx-Spanish dataset are shown in Supplementary Table~\ref{tab: supplemental_MTspanish_results}. We observe that our proposed approach results in a 5.6\,\% absolute reduction in the WER. A further reduction of 4.2\,\% can be achieved by using additional training data.

\begin{table*}[!t]
\centering
\caption{Results on the Multilingual TEDx-Portuguese (MT$_{\text{pt}}$) dataset. `Mean$\pm$Std.' refers to the mean WER over ten runs and the corresponding standard deviation, while `Best' denotes the best (lowest) WER.}
\renewcommand\arraystretch{1.4}
\begin{tabularx}{2\columnwidth}{j v k o y y}
\toprule
Method &Pre-training Set &Training Set &Training Sets Total Size (hours) &Mean$\pm$Std. &Best \\
\midrule
CM-seq2seq~\cite{DBLP:journals/corr/abs-2102-06657} &LRW &CM$_{\text{pt}}$+MT$_{\text{pt}}$ &256 & 70.2$\pm$0.3 & 69.7 \\
\midrule
Ours &LRW &CM$_{\text{pt}}$+MT$_{\text{pt}}$ &256 &\bf 66.0$\pm$0.5 &\bf 65.3 \\
\midrule
Ours &LRW+LRS2+LRS3 &CM$_{\text{pt}}$+MT$_{\text{pt}}$ &917 &\bf 62.4$\pm$0.4 &\bf 62.0 \\
\midrule
Ours &LRW+LRS2+LRS3+AVSpeech &CM$_{\text{pt}}$+MT$_{\text{pt}}$ &1\,558 &\bf 62.1$\pm$0.6 &\bf 61.5 \\
\bottomrule
\end{tabularx}
\label{tab: supplemental_MTportuguese_results}
\end{table*}
\begin{table*}[!t]
\centering
\caption{Results on the CMU-MOSEAS-Portuguese (CM$_{\text{pt}}$) dataset. `Mean$\pm$Std.' refers to the mean WER over ten runs and the corresponding standard deviation, while `Best' denotes the best (lowest) WER.}
\renewcommand\arraystretch{1.4}
\begin{tabularx}{2\columnwidth}{j v k o y y}
\toprule
Method &Pre-training Set &Training Set &Training Sets Total Size (hours) &Mean$\pm$Std. &Best \\
\midrule
CM-seq2seq~\cite{DBLP:journals/corr/abs-2102-06657} &LRW &CM$_{\text{pt}}$+MT$_{\text{pt}}$ &256 & 65.7$\pm$0.5 & 65.4 \\
\midrule
Ours &LRW &CM$_{\text{pt}}$+MT$_{\text{pt}}$ &256 &\bf 57.2$\pm$0.7 &\bf 56.6 \\
\midrule
Ours &LRW+LRS2+LRS3 &CM$_{\text{pt}}$+MT$_{\text{pt}}$ &917 &\bf 53.1$\pm$0.2 &\bf 52.8 \\
\midrule
Ours &LRW+LRS2+LRS3+AVSpeech &CM$_{\text{pt}}$+MT$_{\text{pt}}$ &1\,558 &\bf 51.6$\pm$0.2 &\bf 51.4 \\
\bottomrule
\end{tabularx}
\label{tab: supplemental_CMportuguese_results}
\end{table*}
\begin{table*}[!t]
\centering
\caption{Results on the Multilingual TEDx-French (MT$_{\text{fr}}$) dataset. `Mean$\pm$Std.' refers to the mean WER over ten runs and the corresponding standard deviation, while `Best' denotes the best (lowest) WER.}
\renewcommand\arraystretch{1.4}
\begin{tabularx}{2\columnwidth}{j v k o y y}
\toprule
Method &Pre-training Set &Training Set &Training Sets Total Size (hours) &Mean$\pm$Std. &Best \\
\midrule
CM-seq2seq~\cite{DBLP:journals/corr/abs-2102-06657} &LRW &CM$_{\text{fr}}$+MT$_{\text{fr}}$ &257 & 84.0$\pm$0.7 & 83.2 \\
\midrule
Ours &LRW &CM$_{\text{fr}}$+MT$_{\text{fr}}$ &257 &\bf 74.6$\pm$0.6 &\bf 73.4 \\
\midrule
Ours &LRW+LRS2+LRS3 &CM$_{\text{fr}}$+MT$_{\text{fr}}$ &918 &\bf 67.0$\pm$0.3 &\bf 66.7 \\
\midrule
Ours &LRW+LRS2+LRS3+AVSpeech &CM$_{\text{fr}}$+MT$_{\text{fr}}$ &1\,559 &\bf 67.0$\pm$0.6 &\bf 66.2 \\
\bottomrule
\end{tabularx}
\label{tab: supplemental_MTfrench_results}
\end{table*}
\begin{table*}[!t]
\centering
    \caption{Results on the CMU-MOSEAS-French (CM$_{\text{fr}}$) dataset. `Mean$\pm$Std.' refers to the mean WER over ten runs and the corresponding standard deviation, while `Best' denotes the best (lowest) WER.}
\renewcommand\arraystretch{1.4}
\begin{tabularx}{2\columnwidth}{j v k o y y}
\toprule
Method &Pre-training Set &Training Set &Training Sets Total Size (hours) &Mean$\pm$Std. &Best \\
\midrule
CM-seq2seq~\cite{DBLP:journals/corr/abs-2102-06657} &LRW &CM$_{\text{fr}}$+MT$_{\text{fr}}$ &257 & 79.9$\pm$0.4 & 79.6 \\
\midrule
Ours &LRW &CM$_{\text{fr}}$+MT$_{\text{fr}}$ &257 &\bf 68.4$\pm$0.5 &\bf 67.5 \\
\midrule
Ours &LRW+LRS2+LRS3 &CM$_{\text{fr}}$+MT$_{\text{fr}}$ &918 &\bf 60.1$\pm$0.3 &\bf 59.5 \\
\midrule
Ours &LRW+LRS2+LRS3+AVSpeech &CM$_{\text{fr}}$+MT$_{\text{fr}}$ &1\,559 &\bf 59.1$\pm$0.5 &\bf 58.3 \\
\bottomrule
\end{tabularx}
\label{tab: supplemental_CMfrench_results}
\end{table*}
\begin{table*}[!t]
\caption{Performance (Mean$\pm$Std.) of the pre-trained ASR and VSR Models on the LRS2 dataset. The Baseline VSR model pre-trained on LRW and LRS2 has a mean WER of 33.2$\pm$0.5.}
\renewcommand\arraystretch{1.2}
\begin{tabularx}{2\columnwidth}{f j j j y y y }
\toprule
Method &Training Sets of \\Full Model &Training Sets of \\Pre-trained VSR\\ Model &Training Sets of \\ Pre-trained ASR \\Model & Full Model  &Pre-trained VSR Model &Pre-trained ASR Model \\
\midrule
Ours &LRW+LRS2 &LRW+LRS2 &LRW+LRS2  &\bf 29.5$\pm$0.4 &33.2$\pm$0.5 &3.9$\pm$0.2
\\ \midrule
Ours &LRW+LRS2 &LRW+LRS2 &LRS2 &\bf 30.9$\pm$0.1 &33.2$\pm$0.5 &5.4$\pm$0.1
\\ \midrule
Ours &LRW+LRS2 &LRS2 &LRW+LRS2 &\bf 31.2$\pm$0.4 &52.7$\pm$0.8
&3.9$\pm$0.2
\\ \midrule
Ours &LRW+LRS2 &LRS2 &LRS2 &\bf 33.6$\pm$0.3 &52.7$\pm$0.8
&5.4$\pm$0.1
\\ 
\bottomrule
\end{tabularx}
\label{table: r2_results_on_LRS2}
\end{table*}

\begin{table*}[!t]
\caption{Performance (Mean$\pm$Std.) of the pre-trained ASR and VSR Models on the LRS3 dataset. The Baseline VSR model pre-trained on LRW and LRS3 has a mean WER of 37.8$\pm$0.6.}
\renewcommand\arraystretch{1.2}
\begin{tabularx}{2\columnwidth}{f j j j y y y }
\toprule
Method &Training Sets of \\Full Model &Training Sets of \\Pre-trained VSR\\ Model &Training Sets of \\ Pre-trained ASR \\Model & Full Model  &Pre-trained VSR Model &Pre-trained ASR Model \\
\midrule
Ours &LRW+LRS3 &LRW+LRS3 &LRW+LRS3 &\bf 35.8$\pm$0.5 &37.8$\pm$0.6 &2.2$\pm$0.1
\\\midrule
Ours &LRW+LRS3 &LRW+LRS3 &LRS3 &\bf 36.0$\pm$0.3 &37.8$\pm$0.6 &3.8$\pm$0.1
\\ \midrule
Ours &LRW+LRS3 &LRS3 &LRW+LRS3 &\bf 37.6$\pm$0.3 &75.2$\pm$0.4
&2.2$\pm$0.1
\\ \midrule
Ours &LRW+LRS3 &LRS3 &LRS3 &\bf 37.9$\pm$0.5 &75.2$\pm$0.4
&3.8$\pm$0.1
\\
\bottomrule
\end{tabularx}
\label{table: r2_results_on_LRS3}
\end{table*}

\subsection{Results on Italian}
We manually cleaned the Italian corpus on Multilingual TEDx to exclude videos without visible speakers, resulting in a total of 26387 videos (45.8 hours) for training, 252 videos (0.4 hours) for validation and 309 videos (0.5 hours) for testing. Results on the Multilingual TEDx-Italian dataset  are shown in Supplementary Table~\ref{tab: supplemental_MTitalian_results}. Our proposed approach results in an absolute drop of 5.6\,\%  in the WER. A further reduction of 8\,\% can be achieved by using additional training data.

\subsection{Results on Portuguese}
We manually cleaned the Portuguese corpus on Multilingual TEDx to exclude videos where the speaker is not visible, resulting in a total of 52\,395 videos (81.3 hours) for training, 532 videos (0.7 hours) for validation and 401 videos (0.6 hours) for testing. Results on the Multilingual TEDx-Portuguese dataset are shown in Supplementary Table~\ref{tab: supplemental_MTportuguese_results}. We observe that our proposed approach results in a 4.2\,\% absolute reduction in the WER. A further reduction of 3.9\,\% can be achieved by using additional training data.

We divide the Portuguese corpus on CMU-MOSEAS~\cite{bagher-zadeh-etal-2020-cmu} into 10\,658 videos (17.8 hours) for training and 412 videos (0.7 hours) for testing, respectively. Results on the CMU-MOSEAS-Portuguese dataset are shown in Supplementary Table~\ref{tab: supplemental_CMportuguese_results}. The proposed approach results in a 8.5\,\% absolute reduction in the WER. Using additional training data leads to a further reduction of 5.6\,\%.

\begin{table*}[!t]
\caption{Investigation of the impact of beam size choices on the validation set of the LRS2 dataset.}
\renewcommand\arraystretch{1.05}
\begin{tabularx}{2\columnwidth}{o  k  k  k  k  k  k  k  k}
\toprule
Model &\bf 40 &35 &30 &25 &20 &15 &10 &5  \\ \midrule
Baseline VSR Model &\bf 43.8$\pm$0.3 &43.9$\pm$0.4 &44.0$\pm$0.4 &44.2$\pm$0.5 &44.4$\pm$0.6 &44.6$\pm$0.5 &45.1$\pm$0.5 &46.3$\pm$0.4 \\
\bottomrule
\end{tabularx}
\label{tab: ablation_study_english}
\end{table*}
\begin{table*}[!t]
\caption{Investigation of the impact of beam size choices on the validation set of the CMLR dataset.}
\renewcommand\arraystretch{1.05}
\begin{tabularx}{2\columnwidth}{o  k  k  k  k  k  k  k  k}
\toprule
Model &40 &35 &30 &25 &\bf 20 &15 &10 &5  \\ \midrule
Baseline VSR Model &10.8$\pm$0.10 &10.8$\pm$0.10 &10.8$\pm$0.10 &10.8$\pm$0.08 &\bf 10.8$\pm$0.08 &10.9$\pm$0.06 &10.9$\pm$0.10 &11.3$\pm$0.06 \\
\bottomrule
\end{tabularx}
\label{tab: ablation_study_mandarin}
\end{table*}
\begin{table*}[!t]
\caption{Investigation of the impact of beam size choices on the validation set of the MT$_{\text{es}}$ dataset.}
\renewcommand\arraystretch{1.05}
\begin{tabularx}{2\columnwidth}{o k  k  k  k  k  k  k}
\toprule
Model &\bf 35 &30 &25 &20 &15 &10 &5  \\ \midrule
Baseline VSR Model &\bf 53.9$\pm$0.5 &54.0$\pm$0.3 &54.3$\pm$0.4 &54.7$\pm$0.4 &55.0$\pm$0.4 &55.6$\pm$0.4 &57.2$\pm$0.2 \\
\bottomrule
\end{tabularx}
\label{tab: ablation_study_spanish}
\end{table*}
\begin{table*}[!t]
\caption{Investigation of the impact of beam size choices on the validation set of the MT$_{\text{fr}}$ dataset.}
\renewcommand\arraystretch{1.05}
\begin{tabularx}{2\columnwidth}{o  k  k  k  k  k  k  k  k}
\toprule
Model &\bf 40 &35 &30 &25 &20 &15 &10 &5
\\ \midrule
Baseline VSR Model &\bf 83.1$\pm$0.8 &83.4$\pm$0.9 &83.6$\pm$0.8 &84.2$\pm$0.4 &84.3$\pm$0.7 &85.3$\pm$0.7 &86.5$\pm$1.2 &88.5$\pm$0.6
\\
\bottomrule
\end{tabularx}
\label{tab: ablation_study_french}
\end{table*}
\begin{table*}[!t]
\caption{Investigation of the impact of beam size choices on the validation set of the MT$_{\text{it}}$ dataset.}
\renewcommand\arraystretch{1.05}
\begin{tabularx}{2\columnwidth}{o k  k  k  k  k  k}
\toprule
Model &30 &\bf 25 &20 &15 &10 &5  \\ \midrule
Baseline VSR Model &64.3$\pm$0.7 &\bf 64.2$\pm$0.5 &64.6$\pm$0.8 &65.0$\pm$1.0 &65.5$\pm$0.7 &67.5$\pm$0.7 \\
\bottomrule
\end{tabularx}
\label{tab: ablation_study_italian}
\end{table*}
\begin{table*}[!t]
\caption{Investigation of the impact of beam size choices on the validation set of the MT$_{\text{pt}}$ dataset.}
\renewcommand\arraystretch{1.05}
\begin{tabularx}{2\columnwidth}{o  k  k  k  k  k  k  k  k}
\toprule
Model &40 &\bf 35 &30 &25 &20 &15 &10 &5  \\ \midrule
Baseline VSR Model &68.6$\pm$0.8 &\bf 68.6$\pm$0.8 &68.8$\pm$0.7 &68.9$\pm$0.6 &69.0$\pm$0.6 &69.5$\pm$0.6 &70.1$\pm$0.6 &71.5$\pm$0.6
\\
\bottomrule
\end{tabularx}
\label{tab: ablation_study_portuguese}
\end{table*}

\begin{table*}[!t]
\centering
\caption{Ablation study on the LRS3 dataset. Models are trained on LRW, LRS2, LRS3, and AVSpeech.}
\renewcommand\arraystretch{1.15}
\begin{tabularx}{\textwidth}{u y}
\toprule
Method &WER on LRS3 \\
\midrule\midrule
Our model   & 32.1$\pm$0.3 \\
\midrule
- Audio auxiliary task   &33.2$\pm$0.2 \\
\midrule
- Visual auxiliary task  &32.9$\pm$0.3 \\
\midrule
- Audio auxiliary task, visual auxiliary task &33.6$\pm$0.6 \\
\midrule
- Time masking  &33.2$\pm$0.4 \\
\midrule
- Audio auxiliary task, visual auxiliary task, time masking &33.8$\pm$0.4 \\
\bottomrule
\end{tabularx}
\label{tab: ablation_study_on_lrs3_large_training_sets}
\end{table*}
\begin{table*}[!ht]
\caption{Performance (Mean$\pm$Std.) of the pre-trained ASR and VSR Models on the LRS2 dataset. The baseline  model pre-trained on LRW and LRS2 has a mean WER of 33.2$\pm$0.5. `RSN' and `1D-RSN' refer to the proposed visual and audio front-end modules, respectively. Details are shown in Supplementary Tables 2 and 3, respectively. `SVN' refers to the ShuffleNet v2, where the width multiplier is set to 1. `1D-CNN' refers to the 5-layer CNN module. The detailed architecture of the `1D-CNN' front-end module is presented in Supplementary Table~\ref{net:1dcnn_frontend}.}
\renewcommand\arraystretch{1}
\begin{tabularx}{\linewidth}{l j j y y y }
\toprule
    Method 
    &Encoder of Pre-trained VSR Model
    &Encoder of Pre-trained ASR Model
    &Full Model
    &Pre-trained VSR Model
    &Pre-trained ASR Model \\
\midrule
\midrule
Ours &RSN+Conformer &1D-RSN+Conformer  &\bf 29.5$\pm$0.4
&33.2$\pm$0.5 &3.9$\pm$0.2
\\
\midrule
Ours &SVN+Conformer &1D-RSN+Conformer  &\bf 30.4$\pm$0.5 &37.6$\pm$0.6 &3.9$\pm$0.2
\\
\midrule
Ours &RSN+Conformer &1D-CNN+Conformer  &\bf 31.1$\pm$0.3 &33.2$\pm$0.5 &4.5$\pm$0.2
\\
\midrule
Ours &SVN+Conformer &1D-CNN+Conformer  &\bf 31.4$\pm$0.6 &37.6$\pm$0.6 &4.5$\pm$0.2 \\
\bottomrule
\end{tabularx}
\label{tab: ablation_study_on_premodels_results_lrs2}
\end{table*}

\begin{table*}[!tb]
\caption{The architecture of the 1D-CNN front-end module. The filter shapes are denoted by $\{ \text{Temporal Size}, \text{Channels} \}$ for 1D Convolutional Layers, respectively. The sizes correspond to [Batch Size, Channels, Sequence Length]. $T_a$ denotes the length of audio waveforms.}
\begin{center}{\scalebox{1.0}{
\renewcommand{\arraystretch}{1.6}
\begin{tabular}{c|p{5.5cm}<{\centering}|p{3cm}<{\centering}|p{3cm}<{\centering}}
\hlineB{2}
Component Name
&Layer Type &Input Size &Output Size \\
\hline

\text{Conv}$_1$
&
Conv 1D, 80, 64 &[B, 1, T$_a$] &[B, 64, T$_a//4$]    
\\
\hline

\text{Conv}$_2$
&
Conv 1D, 20, 64 &[B, 64, T$_a//4$] &[B, 64, T$_a//16$]    
\\
\hline

\text{Conv}$_3$
&
Conv 1D, 4, 128 &[B, 64, T$_a//16$] &[B, 128, T$_a//32$]    
\\
\hline

\text{Conv}$_4$
&
Conv 1D, 4, 256 &[B, 128, T$_a//32$] &[B, 256, T$_a//64$]    
\\
\hline

\text{Conv}$_5$
&
Conv 1D, 4, 512 &[B, 256, T$_a//64$] &[B, 512, T$_a//128$]    
\\
\cline{1-4}
Aggregation
&1D Average Pooling, Stride 5
&[B, 512, T$_a//128$] &[B, 512, T$_a//640$]\\
\hlineB{2}
\end{tabular}}}
\end{center}
\label{net:1dcnn_frontend}

\end{table*}

\subsection{Results on French}
We manually cleaned the French corpus on Multilingual TEDx to exclude videos where the speaker is not visible, resulting in a total of 58\,809 videos (84.9 hours) for training, 333 videos (0.4 hours) for validation and 235 videos~(0.3 hours) for testing. Results on the Multilingual TEDx-French dataset are shown in Supplementary Table~\ref{tab: supplemental_MTfrench_results}. The proposed approach results in a 9.4\,\% absolute reduction in the WER. A further reduction of 7.6\,\% can be achieved by using additional training data.

We divide the French corpus on CMU-MOSEAS~\cite{bagher-zadeh-etal-2020-cmu} into 8\,880 videos (15.3 hours) for training and 513 videos~(0.8 hours) for testing, respectively. Results on the CMU-MOSEAS-French dataset are shown in Supplementary Table~\ref{tab: supplemental_CMfrench_results}. We observe that our proposed approach results in a 11.5\,\% absolute reduction in the WER. Furthermore, as expected, the performance is improved by a large margin of 9.3\,\% when additional training data is included.

\subsection{Ablation Study on the Effect of Pre-trained ASR and VSR Models}
In this section, we investigate the impact of pre-trained ASR and VSR models used in the auxiliary tasks. Results on LRS2 are shown in Supplementary Tables~\ref{table: r2_results_on_LRS2} and~\ref{table: r2_results_on_LRS3} below. By replacing the ASR model pre-trained on LRW and LRS2 (WER: 3.9\%) with a model pre-trained only on LRS2 (WER: 5.4\%), we observe that the mean WER increases from 29.5\% to 30.9\%. Similarly by replacing the VSR model pre-trained on LRW+LRS2 (WER: 33.2\%) with a model pre-trained on LRS2 (WER: 52.7\%), the mean WER increases from 29.5\% to 31.2\%. When we use both ASR and VSR models pre-trained on LRS2~(last row of Supplementary Table~\ref{table: r2_results_on_LRS2}), a further increase in the mean WER to 33.6\% is observed, which indicates that a better pre-trained ASR/VSR model leads to improved performance of the full model. Results on LRS3 are reported in Supplementary Table~\ref{table: r2_results_on_LRS3}. In case, when we replace the ASR/VSR model pre-trained on LRW and LRS3 with a model pre-trained on LRS3, the mean WER increases from 35.8\% to 36.0\%/37.6\%. When replacing both ASR and VSR models to LRS3 for initialisation, the mean WER further increases to 37.9\%.

\subsection{Ablation Study on the Effect of Beam Size}
Results on the impact of beam size for multiple languages are presented in Supplementary Tables~\ref{tab: ablation_study_english}, ~\ref{tab: ablation_study_mandarin}, ~\ref{tab: ablation_study_spanish}, \ref{tab: ablation_study_french}, \ref{tab: ablation_study_italian}, and \ref{tab: ablation_study_portuguese}. We optimise the beam size with an interval of 5 based on the validation set. In particular, we have optimised the beam size set to 40 on the English corpus (LRS2 and LRS3), 20 on the Mandarin corpus (CMLR), 35 on the Spanish corpus (CM$_{\text{es}}$ and MT$_{\text{es}}$), 25 on the Italian corpus (MT$_{\text{es}}$), 40 on the French corpus (MT$_{\text{fr}}$) and 35 on the Portuguese corpus (CM$_{\text{pt}}$ and MT$_{\text{pt}}$).

\subsection{Ablation Study on the Effect of Auxiliary Losses when Using a Large Training Set}
Results of the impact of auxiliary losses and time masking on the performance on LRS3 dataset are shown in Supplementary Table~\ref{tab: ablation_study_on_lrs3_large_training_sets}. Note that all models are trained using the LRW, LRS2, LRS3, and AVSpeech datasets, in a total of 1\,459 hours. We observe that overall the results are consistent with the ones presented in Table \ref{tab: ablation_study_on_lrs2_and_lrs3}, i.e. removing either auxiliary loss or training a model without using time masking leads to an increase in the mean WER when compared with the full model. To be specific, by removing a visual auxiliary task results, we observe an absolute increase of 0.8\% in the mean WER. Then, if we also remove the audio auxiliary task, a further increase of 0.7\% in the mean WER is observed. This indicates that training with auxiliary losses can provide a better supervision to the intermediate layer of the model which in turn results in better visual representations and improved performance. Indeed the contribution of the auxiliary losses is smaller when larger sets are used. However, we do believe that this is in line with what we propose in the paper that when don't have access to large training sets then careful design of the model is equally important.

\subsection{Ablation Study on the Effect of Pre-trained VSR Models and ASR Models with Different Architectures}
Results of the impact of the pre-trained VSR and ASR models with different architectures are shown in Supplementary Table~\ref{tab: ablation_study_on_premodels_results_lrs2}. Note that all models are trained using the LRW and LRS2 datasets, in a total of 380 hours. To be specific, replacing the proposed visual/audio front-end modules with the ShuffleNet v2~\cite{ma2018shufflenet} backbone (see Supplementary Table~\ref{net:1dcnn_frontend}) leads to an increase of 4.4\,\% and 0.6\%, respectively, in WER. However, we observe that training a model with auxiliary losses, even when the pre-trained VSR and ASR models have different architectures, outperforms the baseline model. This is in line with what we propose in the paper that training with auxiliary losses can provide a better supervision to the intermediate layer of the model which in turn results in better visual representations and improved performance.

\end{document}